%% file: main.tex
\documentclass[11pt,a4paper]{article}

\usepackage[utf8]{inputenc}
\usepackage[T1]{fontenc}
\usepackage{lmodern}
\usepackage[margin=2.5cm]{geometry}
\usepackage{amsmath,amssymb,amsthm}
\usepackage{bm}
\usepackage{algorithm}
\usepackage{algorithmic}
\usepackage{graphicx}
\graphicspath{{figure/}}
\usepackage{xcolor}
\usepackage[most]{tcolorbox}
\usepackage{tikz}
\usetikzlibrary{shapes.callouts,positioning,shadows,calc,arrows.meta,decorations.pathreplacing,
  patterns,fit,backgrounds,matrix,chains,scopes}
\usepackage{enumitem}
\usepackage{hyperref}
\usepackage[numbers]{natbib}
\usepackage{fontawesome5}
\usepackage{mdframed}
\usepackage{fancyhdr}
\usepackage{etoolbox}
\usepackage{titlesec}
\usepackage{booktabs}
\usepackage{multicol}
\usepackage{caption}
\usepackage{microtype}
\usepackage{listings}
\usepackage{upquote}
\usepackage{accsupp}
\usepackage{pgfplots}
\pgfplotsset{compat=1.18}
\usepackage{subcaption}
\usepackage{colortbl}
\usepackage{multirow}
\usepackage{pifont}
\usepgfplotslibrary{groupplots}
\usepackage{graphicx}
\usepackage{tabularx}
\usepackage{float}
\usepackage{svg}
\usepackage{placeins}

\newcommand{\equalmark}{\textsuperscript{†}}   
\newcommand{\corrmark}{\textsuperscript{‡}}    

\usepackage{pgfplots}
\pgfplotsset{compat=1.18}

\definecolor{lvl1}{RGB}{31,119,180}
\definecolor{lvl2}{RGB}{255,127,14}
\definecolor{lvl3}{RGB}{44,160,44}
\definecolor{lvlA}{RGB}{214,39,40}

\input{glyphtounicode}
\definecolor{titleRed}{HTML}{B71C1C}
\definecolor{theoryBG}{HTML}{FFF8E7}
\definecolor{theoryBorder}{HTML}{D4A843}
\definecolor{practiceBG}{HTML}{E8F4FD}
\definecolor{practiceBorder}{HTML}{3B82B6}
\definecolor{bridgeBG}{HTML}{F0FFF0}
\definecolor{bridgeBorder}{HTML}{2E8B57}
\definecolor{dialogueBG}{HTML}{FAF0FA}
\definecolor{dialogueBorder}{HTML}{8B5E8B}
\definecolor{sharedBG}{HTML}{FAFAFA}
\definecolor{marginJ}{HTML}{B71C1C}
\definecolor{marginH}{HTML}{E65100}
\definecolor{marginS}{HTML}{1565C0}
\definecolor{trackR}{HTML}{B8860B}
\definecolor{trackP}{HTML}{2166AC}
\definecolor{epilogueBG}{HTML}{FFF5F5}
\definecolor{codeBG}{HTML}{1E1E2E}
\definecolor{codeFrame}{HTML}{45475A}
\definecolor{codeGreen}{HTML}{A6E3A1}
\definecolor{codeBlue}{HTML}{89B4FA}
\definecolor{codePurple}{HTML}{CBA6F7}
\definecolor{codeYellow}{HTML}{F9E2AF}
\definecolor{codeRed}{HTML}{F38BA8}
\definecolor{codeOrange}{HTML}{FAB387}
\definecolor{codeComment}{HTML}{6C7086}
\definecolor{warningBG}{HTML}{FFF3E0}
\definecolor{warningBorder}{HTML}{E65100}
\definecolor{insightBG}{HTML}{E8F5E9}
\definecolor{insightBorder}{HTML}{2E7D32}
\definecolor{keyresultBG}{HTML}{FCE4EC}
\definecolor{keyresultBorder}{HTML}{C62828}
\definecolor{pseudoBG}{HTML}{F5F5FF}
\definecolor{pseudoBorder}{HTML}{5C6BC0}
\definecolor{acblue}{RGB}{31,78,121}

\lstdefinestyle{darkpython}{
  language=Python,
  backgroundcolor=\color{codeBG},
  basicstyle=\ttfamily\small\color{white},
  keywordstyle=\color{codePurple}\bfseries,
  stringstyle=\color{codeGreen},
  commentstyle=\color{codeComment}\itshape,
  numberstyle=\tiny\color{codeComment},
  identifierstyle=\color{codeBlue},
  emphstyle=\color{codeYellow},
  emph={[2]self,True,False,None},
  emphstyle={[2]\color{codeOrange}},
  numbers=left, numbersep=8pt,
  frame=single, rulecolor=\color{codeFrame}, framesep=6pt,
  xleftmargin=18pt, framexleftmargin=18pt,
  showstringspaces=false, tabsize=4,
  breaklines=true, breakatwhitespace=true,
  aboveskip=8pt, belowskip=8pt,
  literate={->}{{{\color{codeRed}->}}}2
           {>=}{{{\color{codeRed}>=}}}2
           {<=}{{{\color{codeRed}<=}}}2
           {!=}{{{\color{codeRed}!=}}}2
           {==}{{{\color{codeRed}==}}}2
           {**}{{{\color{codeRed}**}}}2,
}

\lstdefinestyle{darkyaml}{
  basicstyle=\ttfamily\small\color{white},
  backgroundcolor=\color{codeBG},
  keywordstyle=\color{codeYellow},
  stringstyle=\color{codeGreen},
  commentstyle=\color{codeComment}\itshape,
  frame=single, rulecolor=\color{codeFrame}, framesep=6pt,
  xleftmargin=12pt, framexleftmargin=12pt,
  showstringspaces=false, breaklines=true,
  aboveskip=8pt, belowskip=8pt,
  morecomment=[l]{\#},
}

\lstdefinestyle{darkbash}{
  language=bash,
  basicstyle=\ttfamily\small\color{white},
  backgroundcolor=\color{codeBG},
  keywordstyle=\color{codeGreen}\bfseries,
  commentstyle=\color{codeComment}\itshape,
  stringstyle=\color{codeYellow},
  frame=single, rulecolor=\color{codeFrame}, framesep=6pt,
  xleftmargin=12pt, framexleftmargin=12pt,
  showstringspaces=false, breaklines=true,
  aboveskip=8pt, belowskip=8pt,
  literate={~}{{\raise.35ex\hbox{$\scriptstyle\sim$}}}1
           {\$}{{\BeginAccSupp{method=escape,ActualText=}\color{codeGreen}\$\EndAccSupp{}}}1
           {-}{{-}}1,
}

\newtcolorbox{theorytrack}[1][]{%
  enhanced, colback=theoryBG, colframe=theoryBorder, boxrule=1.2pt,
  left=8pt,right=8pt,top=8pt,bottom=8pt,
  fonttitle=\bfseries\sffamily\color{trackR},
  title={\faFlask~\textsf{RESEARCH TRACK}~~#1},
  attach boxed title to top left={xshift=6pt,yshift=-3mm},
  boxed title style={colback=theoryBG,colframe=theoryBorder,boxrule=0.8pt},
  arc=2pt, shadow={1pt}{-1pt}{0pt}{black!15}, breakable
}

\newtcolorbox{practicetrack}[1][]{%
  enhanced, colback=practiceBG, colframe=practiceBorder, boxrule=1.2pt,
  left=8pt,right=8pt,top=8pt,bottom=8pt,
  fonttitle=\bfseries\sffamily\color{trackP},
  title={\faCode~\textsf{PRACTITIONER TRACK}~~#1},
  attach boxed title to top left={xshift=6pt,yshift=-3mm},
  boxed title style={colback=practiceBG,colframe=practiceBorder,boxrule=0.8pt},
  arc=2pt, shadow={1pt}{-1pt}{0pt}{black!15}, breakable
}

\newtcolorbox{bridgebox}[1][]{%
  enhanced, colback=bridgeBG, colframe=bridgeBorder, boxrule=1.5pt,
  left=8pt,right=8pt,top=8pt,bottom=8pt,
  fonttitle=\bfseries\sffamily\color{bridgeBorder},
  title={\faLink~\textsf{BRIDGE}~~#1},
  attach boxed title to top center={yshift=-3mm},
  boxed title style={colback=bridgeBG,colframe=bridgeBorder,boxrule=0.8pt},
  arc=3pt, shadow={1.5pt}{-1.5pt}{0pt}{black!12}, breakable
}

\newtcolorbox{dialogue}[1][]{%
  enhanced, colback=dialogueBG, colframe=dialogueBorder, boxrule=1.5pt,
  left=10pt,right=10pt,top=10pt,bottom=10pt,
  fonttitle=\bfseries\sffamily\large\color{dialogueBorder},
  title={\faTheaterMasks~\textsf{THE LOBBY}~~#1},
  attach boxed title to top center={yshift=-3mm},
  boxed title style={colback=dialogueBG,colframe=dialogueBorder,boxrule=0.8pt},
  arc=4pt, shadow={2pt}{-2pt}{0pt}{black!15}, breakable
}

\newtcolorbox{epilogue}[1][]{%
  enhanced, colback=epilogueBG, colframe=dialogueBorder!60, boxrule=1pt,
  left=10pt,right=10pt,top=8pt,bottom=8pt,
  fonttitle=\bfseries\sffamily\color{dialogueBorder!80},
  title={\faStar~\textsf{EPILOGUE}~~#1},
  attach boxed title to top center={yshift=-3mm},
  boxed title style={colback=epilogueBG,colframe=dialogueBorder!60,boxrule=0.8pt},
  arc=3pt, breakable
}

\newtcolorbox{chaptermap}{%
  enhanced, colback=sharedBG, colframe=black!40, boxrule=1pt,
  left=8pt,right=8pt,top=8pt,bottom=8pt,
  fonttitle=\bfseries\sffamily,
  title={\faMap~\textsf{READING MAP}},
  attach boxed title to top left={xshift=6pt,yshift=-3mm},
  boxed title style={colback=sharedBG,colframe=black!40,boxrule=0.8pt},
  arc=2pt, breakable
}

\newtcolorbox{warningbox}[1][]{%
  enhanced, colback=warningBG, colframe=warningBorder, boxrule=1pt,
  left=8pt,right=8pt,top=6pt,bottom=6pt,
  fonttitle=\bfseries\sffamily\color{warningBorder},
  title={\faExclamationTriangle~\textsf{WARNING}~~#1},
  attach boxed title to top left={xshift=6pt,yshift=-3mm},
  boxed title style={colback=warningBG,colframe=warningBorder,boxrule=0.8pt},
  arc=2pt, breakable
}

\newtcolorbox{insightbox}[1][]{%
  enhanced, colback=insightBG, colframe=insightBorder, boxrule=1pt,
  left=8pt,right=8pt,top=6pt,bottom=6pt,
  fonttitle=\bfseries\sffamily\color{insightBorder},
  title={\faLightbulb~\textsf{KEY INSIGHT}~~#1},
  attach boxed title to top left={xshift=6pt,yshift=-3mm},
  boxed title style={colback=insightBG,colframe=insightBorder,boxrule=0.8pt},
  arc=2pt, breakable
}

\newtcolorbox{keyresult}[1][]{%
  enhanced, colback=keyresultBG, colframe=keyresultBorder, boxrule=1.2pt,
  left=8pt,right=8pt,top=6pt,bottom=6pt,
  fonttitle=\bfseries\sffamily\color{keyresultBorder},
  title={\faTrophy~\textsf{KEY RESULT}~~#1},
  attach boxed title to top left={xshift=6pt,yshift=-3mm},
  boxed title style={colback=keyresultBG,colframe=keyresultBorder,boxrule=0.8pt},
  arc=2pt, breakable
}

\newtcolorbox{pseudobox}[1][]{%
  enhanced, colback=pseudoBG, colframe=pseudoBorder, boxrule=1pt,
  left=8pt,right=8pt,top=6pt,bottom=6pt,
  fonttitle=\bfseries\sffamily\color{pseudoBorder},
  title={\faCogs~\textsf{ALGORITHM}~~#1},
  attach boxed title to top left={xshift=6pt,yshift=-3mm},
  boxed title style={colback=pseudoBG,colframe=pseudoBorder,boxrule=0.8pt},
  arc=2pt, breakable
}

\newtcolorbox{sharedtrack}[1][]{%
  enhanced, colback=sharedBG, colframe=black!40, boxrule=1.2pt,
  left=8pt,right=8pt,top=8pt,bottom=8pt,
  fonttitle=\bfseries\sffamily\color{black!70},
  title={\faLayerGroup~\textsf{SHARED TRACK}~~#1},
  attach boxed title to top left={xshift=6pt,yshift=-3mm},
  boxed title style={colback=sharedBG,colframe=black!40,boxrule=0.8pt},
  arc=2pt, shadow={1pt}{-1pt}{0pt}{black!12}, breakable
}

\newtheorem{proof-env}{Proof}
\newtheorem{definition}{Definition}

\usepackage{tcolorbox}
\usepackage{enumitem}
\usepackage{caption}

\tcbset{
  casebox/.style={
    colback=#1,
    colframe=gray!50,
    fonttitle=\bfseries\small,
    boxrule=0.5pt,
    arc=2pt,
    left=6pt, right=6pt, top=4pt, bottom=4pt,
    before skip=6pt, after skip=6pt,
  }
}

\fancypagestyle{empty}{%
  \fancyhf{}%
}
\fancypagestyle{plain}{%
  \fancyhf{}%
}

\begin{document}
\thispagestyle{empty}
{
\centering
\vspace*{-0.5cm}

{\LARGE\bfseries\sffamily\color{titleRed} Web2BigTable: A Bi-Level Multi-Agent LLM System for Internet-Scale Information Search and Extraction}

\vspace{25pt}

{\small
Yuxuan~Huang$^{1}$\equalmark,\;
Yihang~Chen$^{3}$\equalmark,\;
Zhiyuan~He$^{3}$,\;
Yuxiang~Chen$^{3}$,\;
Ka Yiu~Lee$^{2}$,\;
Huichi~Zhou$^{3}$,\;
Weilin~Luo$^{2}$,\;
Meng~Fang$^{1}$,\;
and Jun~Wang$^{3}$\corrmark\\[6pt]
{\footnotesize $^1$University of Liverpool\quad
$^2$Huawei Noah's Ark Lab\quad
$^3$University College London}
}

\par
}
\vspace{15pt}

\begin{abstract}
Agentic web search increasingly faces two distinct demands: deep reasoning over a single target, and structured aggregation across many entities and heterogeneous sources. Current systems struggle on both fronts. Breadth-oriented tasks demand schema-aligned outputs with wide coverage and cross-entity consistency, while depth-oriented tasks require coherent reasoning over long, branching search trajectories. We introduce \textbf{Web2BigTable}, a multi-agent framework for web-to-table search that supports both regimes. Web2BigTable adopts a bi-level architecture in which an upper-level orchestrator decomposes the task into sub-problems and lower-level worker agents solve them in parallel. Through a closed-loop run--verify--reflect process, the framework jointly improves decomposition and execution over time via persistent, human-readable external memory, with self-evolving updates to each single-agent. During execution, workers coordinate through a shared workspace that makes partial findings visible, allowing them to reduce redundant exploration, reconcile conflicting evidence, and adapt to emerging coverage gaps. Web2BigTable sets a new state of the art on WideSearch, reaching an Avg@4 Success Rate of \textbf{38.50} ($7.5\times$ the second best at 5.10), Row F1 of \textbf{63.53} (+25.03 over the second best), and Item F1 of \textbf{80.12} (+14.42 over the second best). It also generalises to depth-oriented search on XBench-DeepSearch, achieving 73.0 accuracy. Code is available at \url{https://github.com/web2bigtable/web2bigtable}.


\end{abstract}

\begin{figure}[H]
    \centering
    \includegraphics[width=0.90\linewidth]{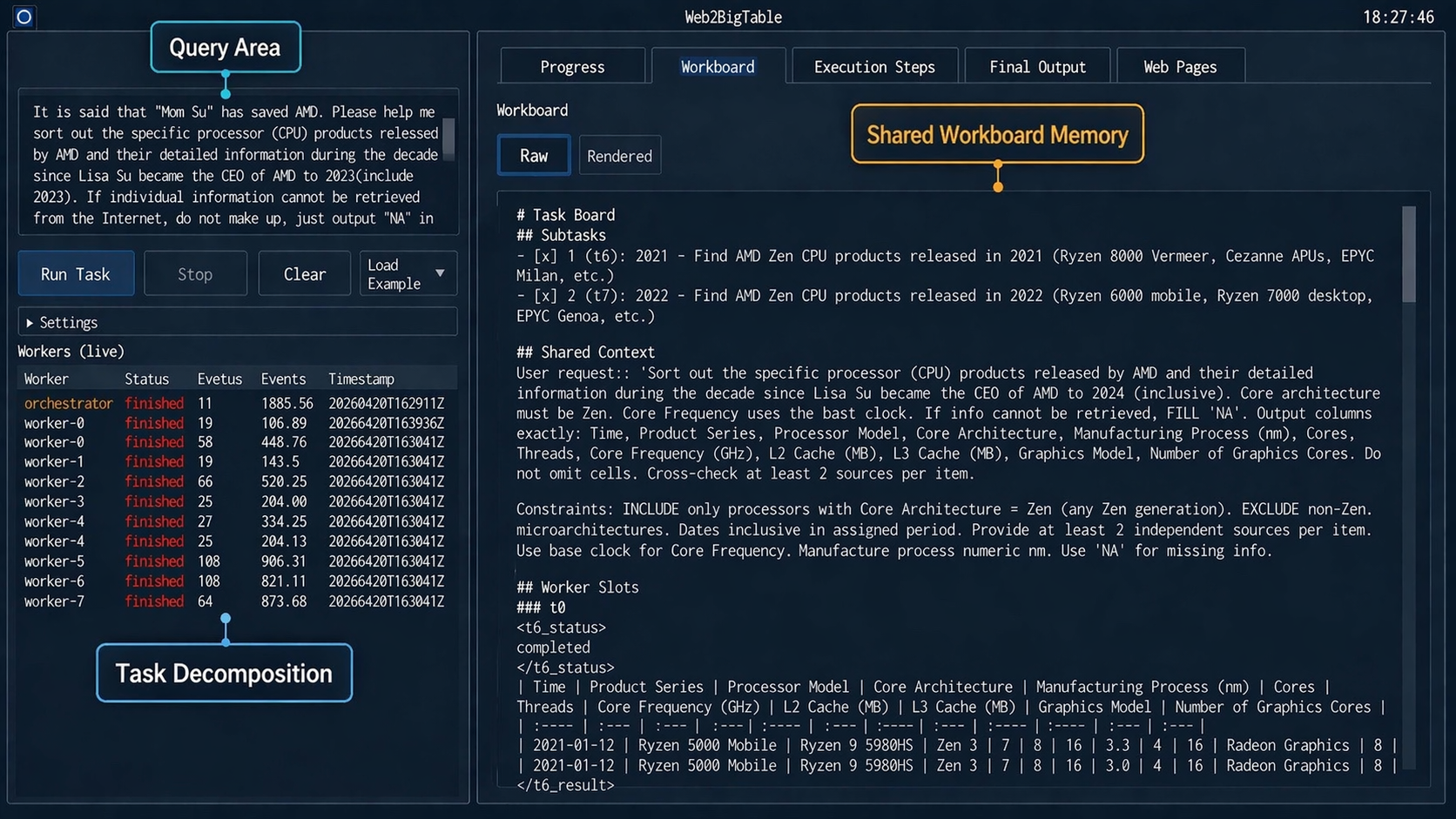}
    \caption{Web2BigTable interface during query execution. The left panel displays the user query (top left), the task decomposition status indicating the orchestrator has partitioned the query into parallel subtasks (bottom left). The right panel shows the shared workboard memory in raw Markdown view, organised into three sections: a subtask checklist tracking each worker's assignment and completion status, a shared context block containing the extraction constraints and target schema, and individual worker result slots where each worker writes its structured table output into a dedicated tag-partitioned region.}

    \label{fig:tui}
\end{figure}

\newpage

\input{sec/intro}

\input{sec/formulation-and-method}

\input{sec/experiment}
\input{sec/related}

\input{sec/conclusion}

\bibliographystyle{plainnat} 
\bibliography{sec/main}

\appendix
\input{sec/appendix}
\end{document}

%% file: sec/intro.tex
\section{Introduction}
\label{sec:intro}

Searching the open web and organising its contents into structured outputs is an important capability for LLM agents operating over real-world information~\citep{brach2026scrapegraphai}. Agentic web search broadly spans two regimes. In \emph{deep search}, agents iteratively retrieve, read, and reason to resolve a single complex query~\citep{li2025webthinker, jin2025search}. In \emph{wide search}, the goal is instead to assemble a consistent, structured view over many entities and attributes grounded in heterogeneous sources. A recent formulation of these regimes is web-to-table construction, in which an agent, given a natural-language request, must autonomously search the web and return a schema-aligned table whose rows correspond to entities and whose columns capture the requested attributes~\citep{wong2025widesearch}. While both regimes require tool use and multi-step reasoning, wide search introduces stronger demands on coverage, decomposition, coordination, and aggregation at scale, making it a challenging and practically important setting for evaluating general web agents.

Despite recent progress, existing systems remain limited on both axes. To see the problem, compare two regimes. A query from XBench-DeepSearch~\citep{chen2025xbench} such as \textit{a member of a fifth-generation Korean girl group has a brother 12 years younger; which group is it?} requires chaining indirect clues across entertainment databases, fan wikis, and biographical records to identify a single entity, which is deep search. A query from WideSearch~\citep{wong2025widesearch} such as \textit{list every Taylor Swift concert from 2010 to 2025 with date, city, and venue} returns hundreds of rows, each of which must be correct and mutually consistent, which is wide search. The first demands long, coherent reasoning; the second demands broad, verified coverage. A monolithic agent handles neither well at scale: the context saturates, errors compound, and the fixed plan made at the start cannot adapt to what later steps uncover. Hierarchical agent frameworks~\citep{qian2025macnet,lee2026infoseeker,zhou2025memento} and automatically searched workflow pipelines~\citep{zhang2025aflow, hu2025adas} partially address this issue through task decomposition, but often rely on fixed or weakly adaptive planning strategies, with limited feedback from downstream execution to upstream decomposition~\citep{liu2025selectdecompose, yu2026adaptorch}. Meanwhile, recent self-evolving and memory-augmented agents improve execution through reusable external skills~\citep{ge2025samule, wu2025evolver, sage2025}
or structured long-term memory~\citep{amem2025, memoryr1_2025}, yet this adaptation has largely been studied at a single level, without jointly refining how tasks are decomposed and how sub-tasks are solved. As a result, effective agentic web search requires not only decomposition, but also a mechanism to co-adapt planning and execution while enabling multiple workers to coordinate, avoid redundant exploration, reconcile conflicting evidence, and expose coverage gaps during search.

\renewcommand{\thedefinition}{}
\begin{definition}[Web-to-Table Search]
\textit{Given a natural-language query and a target schema, produce a structured table by searching the open web, where each row is a distinct entity, each column is a requested attribute, and every cell is independently verified against web sources.}
\end{definition}

To address these challenges, we propose \textbf{Web2BigTable}, a multi-agent framework for web-to-table search that supports both breadth-oriented and depth-oriented instances of this task. Web2BigTable uses a bi-level architecture in which an upper-level orchestrator decomposes the task into sub-problems and lower-level worker agents solve them in parallel. The framework improves over time through a closed-loop run-verify-reflect process that jointly refines decomposition and execution, making it self-evolving through persistent, editable external memory. Concretely, the orchestrator accumulates human-readable decomposition skills that improve future task partitioning, while workers acquire reusable execution skills for retrieval, evidence verification, and intermediate synthesis. All adaptation is mediated through external memory rather than gradient updates, leaving the underlying LLMs frozen throughout. During execution, workers coordinate through a shared workspace that makes partial findings visible, enabling them to reduce redundant exploration, resolve conflicts, and adapt their search trajectories to the evolving global state.
  
We evaluate Web2BigTable on two  benchmarks covering both breadth-oriented and depth-oriented web search settings. On WideSearch~\citep{wong2025widesearch}, a benchmark for broad-coverage search, Web2BigTable achieves state-of-the-art performance, reaching an Avg@4 Success Rate of \textbf{38.50} ($7.5\times$ the second best at 5.10), a Row F1 of \textbf{63.53} (+25.03 over the second best), and an Item F1 of \textbf{80.12} (+14.42 over the second best). The framework also generalises to depth-oriented search, achieving 73.0 accuracy on XBench-DeepSearch~\citep{chen2025xbench}.
 
Our main contributions are as follows:
\begin{itemize}
    \item We introduce \textbf{Web2BigTable}, a multi-agent framework for web-to-table search that supports both breadth-oriented and depth-oriented search regimes.
    \item We propose a bi-level adaptation mechanism that jointly evolves task decomposition and worker execution through persistent, human-readable external memory, with self-evolving to the underlying LLMs.
    \item We develop an asynchronous coordination mechanism that enables parallel workers to share progress, reduce redundant exploration, reconcile conflicting evidence, and respond to emerging coverage gaps during search.
    \item We achieve state-of-the-art results on WideSearch and demonstrate strong generalisation to the structurally distinct XBench-DeepSearch benchmark.
\end{itemize}

%% file: sec/formulation-and-method.tex
\begin{figure}[t]
  \centering
  \includegraphics[width=.9\linewidth]{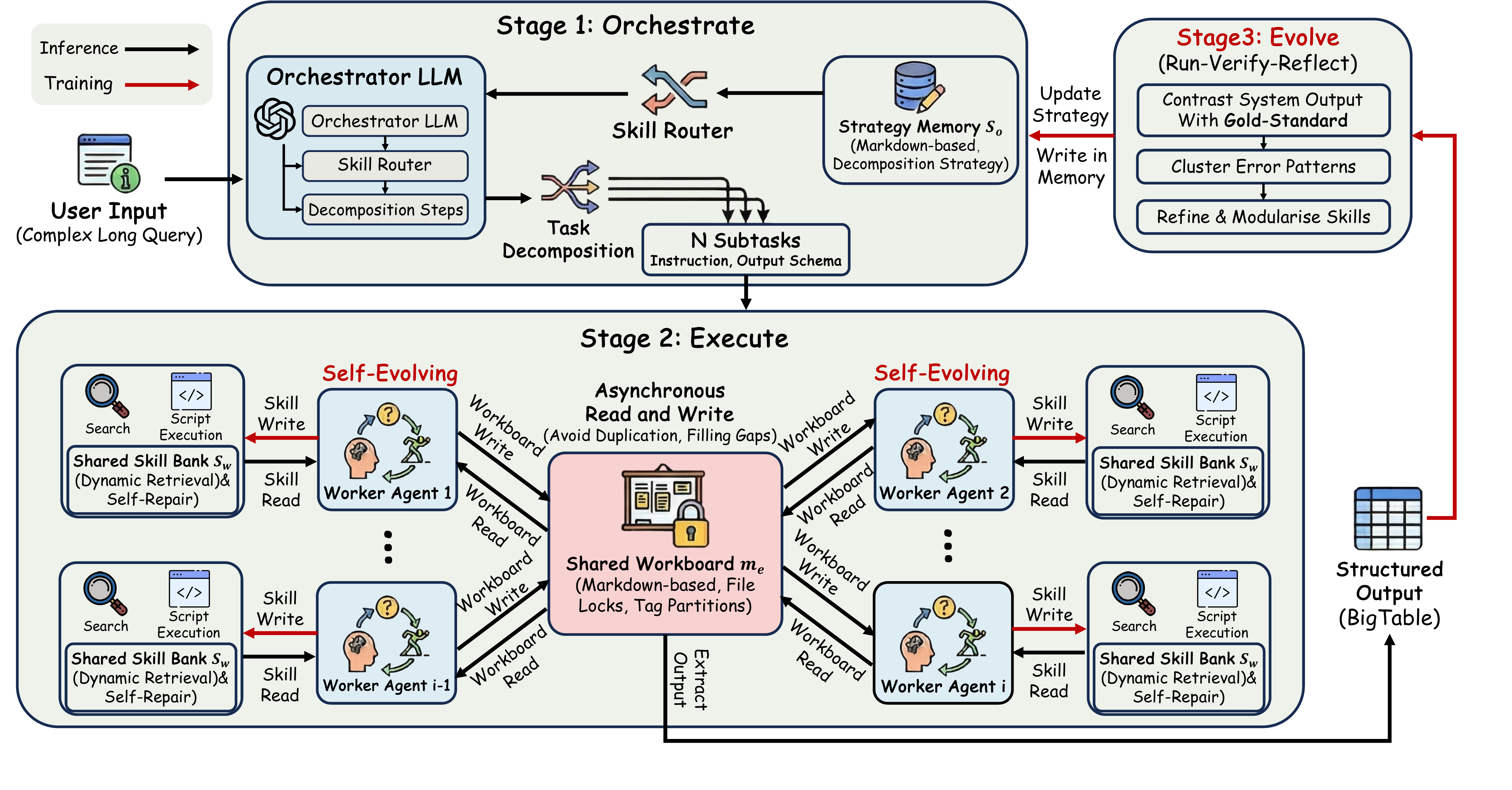}
  \caption{Architecture of \textbf{Web2BigTable}, a unified training and inference framework. The upper-level orchestrator decomposes each query into subtasks using the Orchestrator Skills $\mathcal{S}_o$, and lower-level workers execute them in parallel, drawing execution skills from the shared Worker Skills $\mathcal{S}_w$ and coordinating asynchronously through the Workboard $m_e$, protected by file locks and tag partitioning. Red arrows denote the additional write-back and update flows that are active only at training time, where a verify--reflect step distils each episode's trajectories into monotone updates to $\mathcal{S}_o$ and $\mathcal{S}_w$. At inference the red paths are inactive and the two skill banks are consumed read-only. Both phases operate entirely through external memory, leaving the underlying LLMs frozen throughout.}

  \label{fig:team_arch}
\end{figure}

\section{Web2BigTable: A Bi-Level Memory-Based Framework}
\label{sec:method}

Web2BigTable instantiates the web-to-table search task as a concrete multi-agent framework organised around two design principles. The first is bi-level self-evolving, which jointly refines how queries are decomposed and how sub-tasks are executed by learning two persistent skill banks (Section~\ref{sec:overview}) through an automated training pipeline (Section~\ref{sec:self-evolving}). The second is multi-agent coordination via a shared workboard, which lets concurrent workers observe others' progress, avoid redundant exploration, and reconcile conflicting evidence (Section~\ref{sec:execution}). Both features operate entirely over external memory, leaving the underlying LLMs frozen throughout.

\subsection{Problem Definition}
\label{sec:problem-definition}

Web-to-table search task poses a qualitatively different challenge from conventional agentic web search. Whereas deep question answering converges on a single free-form response, web-to-table search demands a structured table whose rows enumerate many entities and whose cells are each anchored in live web evidence. The task is therefore inherently \emph{wide}: a single query may span hundreds of rows, and success is governed by breadth of entity coverage rather than by depth of reasoning on any individual target.

Formally, an instance is a pair $\mathcal{T} = \langle q, \mathcal{W} \rangle$. The user query $q$ specifies a schema of attributes to extract, for instance columns \textit{Date, Venue, City} across hundreds of concert entries and $\mathcal{W}$ denotes the open web environment that the agent queries through retrieval and browsing tools. A valid output is a table $X \in \mathcal{X}$ whose rows enumerate the entities in $q$, whose columns instantiate the attributes, and whose cells each hold a value verified against web sources. Throughout this paper we adopt a scalar global utility $U(X) \in [0, 1]$ summarising overall extraction quality as the optimisation target.

Any policy $\pi$ solving this task unfolds as an action--observation loop. At each step $t$, $\pi$ samples the next action conditioned on the query $q$ and the accumulated interaction history $h^t = (o^1, x^1, \dots, o^t, x^t)$:
\begin{equation}
  x^{t+1} \sim \pi(\cdot \mid q, h^t).
  \label{eq:single-agent-policy}
\end{equation}
Each action is either a \emph{tool-call} such as a search query or file operation that queries $\mathcal{W}$ and returns an observation $o^{t+1}$, the retrieved content such as a page, snippet, or tool output, appended to $h^{t+1}$, or a \emph{terminal} action that emits the completed table as the policy's final response and ends the trajectory at step $T$. The output table is precisely $X := x^T$, the content of this terminal action.
\subsection{System Architecture}
\label{sec:overview}

The architecture of Web2BigTable is shaped by two structural demands of web-to-table search. First, a single instance spans many entities under a structured schema. The single-agent policy of Equation~\eqref{eq:single-agent-policy} must therefore handle retrieval, state tracking, and synthesis across hundreds of rows within a bounded context window, while remaining unable to exploit the conditional independence across entities that the task naturally exhibits~\citep{li2025webthinker}. This motivates a factorisation of the policy into two levels: an upper-level orchestrator that decomposes the query into independent subtasks, and a lower-level pool of workers that resolves these subtasks in parallel, with the two layers strictly separated and interacting only through shared memory.

Second, the bi-level system must sustain state at two temporal scales. Concurrent workers need to exchange partial evidence within a query to avoid redundant exploration and to reconcile conflicting retrievals, and the system as a whole needs to retain the decomposition and execution skills it acquires across queries. Since the underlying language models remain frozen, such state cannot be absorbed into a single context window; it is instead externalised into two complementary memory regimes:
\begin{itemize}
  \item \textbf{Long-term semantic memory}: a persistent store of skills that evolves only during training and is frozen at inference. It spans two levels: the upper-level orchestrator skills $\mathcal{S}_o$, containing decomposition strategies used by the orchestrator, and the lower-level worker skills $\mathcal{S}_w$, holding execution skills shared across all workers for information seeking.
  \item \textbf{Short-term working memory}: a scratchpad that is transient within a single episode, realised by the workboard $m_e$, through which concurrent workers coordinate their ongoing trajectories.
\end{itemize}

Together, these memories replace the single-agent policy of Equation~\eqref{eq:single-agent-policy} with a bi-level factorisation,
\begin{equation}
\begin{gathered}
  \boldsymbol{\tau} = (\tau_1, \dots, \tau_N) \sim \pi_o(\cdot \mid q, \mathcal{S}_o), \\[2pt]
  x_i \sim \pi_w^{(i)}(\cdot \mid \tau_i, m_e, s_i), \quad s_i \in \mathcal{S}_w,
\end{gathered}
\label{eq:bilevel-policy}
\end{equation}
where each worker's execution skill $s_i$ is retrieved from the shared bank $\mathcal{S}_w$ via the Memento-Skills mechanism~\citep{zhou2026memento}; the orchestrator selects a single decomposition skill from $\mathcal{S}_o$ through a task-router keyed on the query's structural type (Section~\ref{sec:orchestration}).
Equation~\eqref{eq:bilevel-policy} recasts the single-agent policy of Equation~\eqref{eq:single-agent-policy} through three simultaneous decompositions. First, the monolithic $\pi$ splits into an upper-level orchestrator policy $\pi_o$ and a pool of lower-level worker policies $\{\pi_w^{(i)}\}_{i=1}^{N}$: the orchestrator commits to a decomposition $\boldsymbol{\tau} = (\tau_1, \dots, \tau_N)$ under the orchestrator skills $\mathcal{S}_o$, and each worker $i$ independently produces a partial output $x_i$ conditioned on the shared workboard $m_e$ and a retrieved skill $s_i \in \mathcal{S}_w$. Second, the global interaction history $h^t$ that a single agent would otherwise accumulate within its context window is externalised into the shared workboard $m_e$ and partitioned across workers as $\{h_i^t\}_i$; each worker reads only the slice of $m_e$ relevant to its subtask, so the per-step context cost depends on subtask complexity rather than on the full table size. Third, the output $X$ of Section~\ref{sec:problem-definition} is assembled from the partial outputs as $X = (x_1, \dots, x_N)$, and the global utility $U(X)$ becomes the optimisation target used to evolve $\mathcal{S}_o$ and $\mathcal{S}_w$ during training in Section~\ref{sec:self-evolving}. The two timescales are naturally separated: $m_e$ turns over within minutes, whilst $\mathcal{S}_o$ and $\mathcal{S}_w$ accumulate experience across queries over hours or days.

The workboard $m_e$ is not merely a message relay but a shared epistemic state. Because every worker can read the entire workboard whilst writing only to its own tagged region, the architecture makes partial findings immediately available and keeps writes non-destructive over disjoint slots, yielding a form of asynchronous consensus that scales with the worker pool whilst preserving the simplicity of a plain Markdown document. The specific adaptive behaviours this design enables are detailed in Section~\ref{sec:execution}.

\paragraph{Memory updates.}
The three memory regimes evolve at different timescales through dedicated update operators. The workboard is refreshed at every inner step $t$ as contributions from the active worker set $\mathcal{A}^t$ are merged,
\begin{equation}
  m_e^{t+1} = \mathcal{M}_e\big(m_e^t, \{h_i^{t+1}\}_{i \in \mathcal{A}^t}\big),
  \label{eq:execution-memory-update}
\end{equation}
with per-worker writes serialised through file locks so that no two updates collide on disjoint slots. As a short-term working memory, $m_e$ stores the ongoing natural-language trace of a single episode, holding partial findings and intermediate observations that workers read and extend as the query unfolds, and the entire document is discarded once the episode terminates. The two skill banks evolve only across training episodes, through parallel reflect operators,
\begin{align}
  \mathcal{S}_o^{k+1} &= \mathcal{M}_o\big(\mathcal{S}_o^k, r_o^{k+1}\big), \label{eq:strategy-memory-update} \\
  \mathcal{S}_w^{k+1} &= \mathcal{M}_w\big(\mathcal{S}_w^k, r_o^{k+1}\big), \label{eq:skill-bank-update}
\end{align}
where $r_o^{k+1}$ is the structured error report distilled from episode $k$'s trajectories against its gold reference. Both $\mathcal{M}_o$ and $\mathcal{M}_w$ are realised as LLM-driven Run--Verify--Reflect pipelines (Section~\ref{sec:self-evolving}) that append new skills monotonically, without fine-tuning the underlying LLMs. By contrast, $\mathcal{S}_o$ and $\mathcal{S}_w$ constitute a long-term semantic memory, stored as human-readable \texttt{SKILL.md} files that crystallise the distilled, generalised procedures extracted from completed episodes and accumulate monotonically across runs. The two memory regimes are therefore complementary rather than redundant: the workboard records what the system is currently doing and is thrown away, whereas the skill banks record what the system has learned to do and are kept.

\paragraph{Two-phase pipeline: training and inference.}
The split between long-term skill banks and short-term workboard partitions Web2BigTable's operation into two cleanly decoupled phases, separating how skills are accumulated from how they are exploited. The \emph{training} phase (Section~\ref{sec:self-evolving}) runs on a small set of queries paired with gold-standard tables, distilling each episode's outcome into updates of $\mathcal{S}_o$ and $\mathcal{S}_w$ through a run--verify--reflect loop. The \emph{inference} phase (Section~\ref{sec:inference}) freezes the evolved banks $(\mathcal{S}_o^{*}, \mathcal{S}_w^{*})$ and answers unseen user queries in a single forward pass, with no further learning. We present the two phases in their natural order: training first, to specify how the banks are acquired, followed by inference, which consumes them unchanged.

\subsection{Training Phase: Self-Evolving the Skill Banks}
\label{sec:self-evolving}

The policy in Equation~\eqref{eq:bilevel-policy} is only as good as the skills populating $\mathcal{S}_o$ and $\mathcal{S}_w$. Web2BigTable acquires both banks autonomously through a dedicated training phase that runs on a small split of queries paired with gold-standard tables, so that extraction quality can be evaluated objectively. At the end of training the evolved banks $(\mathcal{S}_o^{*}, \mathcal{S}_w^{*})$ are frozen and consumed unchanged by the inference phase of Section~\ref{sec:inference}.

\paragraph{Training objective.}
Training maximises the expected global utility $U(X) = \text{Item-F1}(X,\, X^{\text{gold}})$, a cell-level scoring function that compares the predicted table $X$ against the gold reference $X^{\text{gold}}$ using type-specific comparators (exact match, numeric tolerance, URL normalisation, and LLM-based semantic judgement for free text). Because subtasks are scoped to disjoint entity partitions, each worker contributes additively to $U$: improvements to $\mathcal{S}_o$ (yielding better decompositions) and to $\mathcal{S}_w$ (yielding better per-worker execution) both raise $U$ without interfering, which justifies training the two skill banks jointly rather than sequentially.

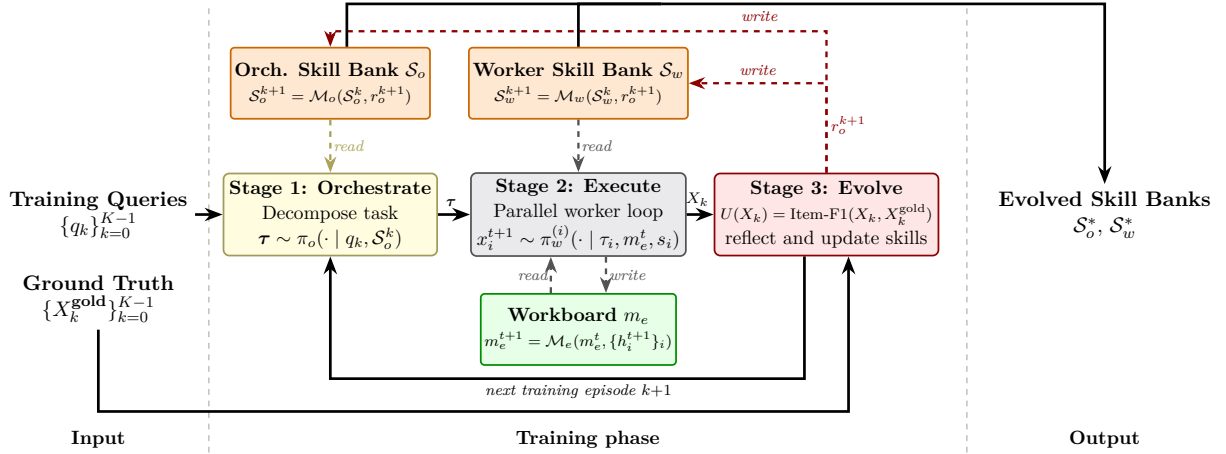
\begin{figure*}[t]
\centering
\resizebox{\linewidth}{!}{%
\begin{tikzpicture}[
  >=Stealth, font=\small,
  stage/.style={rounded corners=3pt, line width=0.9pt, align=center,
                minimum height=1.5cm, minimum width=3.2cm, inner sep=3pt, draw},
  s1/.style={stage, fill=yellow!15,      draw=yellow!55!black},
  s2/.style={stage, fill=blue!8!gray!15, draw=gray!65!black},
  s3/.style={stage, fill=red!10,         draw=red!55!black},
  mem/.style={rounded corners=2pt, line width=0.9pt, align=center,
              minimum height=1.3cm, minimum width=3.0cm, inner sep=2pt, draw},
  mem_e/.style={mem, fill=green!10,   draw=green!55!black},
  mem_bo/.style={mem, fill=orange!18, draw=orange!80!black},
  io/.style={rounded corners=3pt, line width=1.2pt, align=center,
             minimum height=1.2cm, minimum width=2.5cm, inner sep=3pt,
             draw=violet!70!black, fill=violet!12, font=\bfseries},
  eq/.style={font=\scriptsize\itshape, inner sep=1pt},
  mainflow/.style={->, line width=1.4pt},
  link1/.style={->, line width=1.2pt, yellow!60!black, dashed},
  link2/.style={->, line width=1.2pt, gray!65!black,   dashed},
  link3/.style={->, line width=1.2pt, red!55!black,    dashed},
  loop/.style={->, line width=1.4pt}
]

\node[s1] (S1) at (0, 0)
  {\textbf{Stage 1: Orchestrate}\\[1pt] Decompose task\\[1pt]
   $\boldsymbol{\tau} \sim \pi_o(\cdot \mid q_k, \mathcal{S}_o^k)$};
\node[s2] (S2) at (4.5, 0)
  {\textbf{Stage 2: Execute}\\[1pt] Parallel worker loop\\[1pt]
   $x_i^{t+1} \sim \pi_w^{(i)}(\cdot \mid \tau_i, m_e^t, s_i)$};
\node[s3] (S3) at (9.0, 0)
  {\textbf{Stage 3: Evolve}\\[1pt]
   {\scriptsize $U(X_k) = \text{Item-F1}(X_k, X_k^{\text{gold}})$}\\[1pt]
   reflect and update skills};

\node[mem_bo] (mo) at (0, 2.4)
  {\textbf{Orch. Skill Bank} $\mathcal{S}_o$\\[1pt]
   {\scriptsize $\mathcal{S}_o^{k+1} = \mathcal{M}_o(\mathcal{S}_o^{k}, r_o^{k+1})$}};
\path let \p1=($(S2.east)-(S2.west)$) in
  node[mem_bo, minimum width=\x1] (B) at (4.5, 2.4)
    {\textbf{Worker Skill Bank} $\mathcal{S}_w$\\[1pt]
     {\scriptsize $\mathcal{S}_w^{k+1} = \mathcal{M}_w(\mathcal{S}_w^{k}, r_o^{k+1})$}};

\node[mem_e] (me) at (4.5, -2.1)
  {\textbf{Workboard} $m_e$\\[1pt]
   {\scriptsize $m_e^{t+1} = \mathcal{M}_e(m_e^t, \{h_i^{t+1}\}_i)$}};

\node[font=\bfseries, align=center, left=0.5cm of S1] (in)
  {Training Queries\\$\{q_k\}_{k=0}^{K-1}$};
\draw[mainflow] (in.east) -- node[eq, above]{ } (S1.west);

\node[font=\bfseries, align=center, below=0.4cm of in] (inGT)
  {Ground Truth\\$\{X_k^{\text{gold}}\}_{k=0}^{K-1}$};
\draw[mainflow] (inGT.south) |- (9.4, -3.6) -- ([xshift=0.4cm]S3.south);
\node[eq, anchor=south, yshift=1pt] at (5.0, -3.6) { };


\draw[mainflow] (S1.east) -- node[eq, above,xshift=-0.05cm,yshift=0.1cm]{$\boldsymbol{\tau}$} (S2.west);
\draw[mainflow] (S2.east) -- node[eq, above,xshift=-0.05cm,yshift=0.1cm]{$X_k$} (S3.west);

\draw[loop] ([xshift=-0.4cm]S3.south) -- (8.6, -3.0) -- (0, -3.0) -- (S1.south);
\node[eq, anchor=north, yshift=-2pt] at (4.5, -3.0) {next training episode $k{+}1$};

\draw[link1] (mo.south) -- node[eq, midway, right]{read} (S1.north);
\draw[link2] (B.south)  -- node[eq, midway, right]{read} (S2.north);

\draw[link3] (S3.north) |- node[eq, pos=0.75, above, yshift=0.1cm]{write} (B.east);
\draw[link3] (S3.north) -- (9, 3.35) -- (0, 3.35) -- (mo.north);
\node[eq, anchor=south, yshift=0.1cm, text=red!55!black] at (7.78, 3.40) {write};
\node[eq, anchor=south, yshift=1pt, text=red!55!black] at (9.4, 1.4) {$r_o^{k+1}$};

\draw[link2] ([xshift=-0.5cm]me.north) -- node[eq, midway, left]{read} ([xshift=-0.5cm]S2.south);
\draw[link2] ([xshift=0.5cm]S2.south) -- node[eq, midway, right]{write} ([xshift=0.5cm]me.north);

\begin{scope}[on background layer]
  \coordinate (sepL) at ($(in.east)!0.5!(S1.west)$);
  \coordinate (sepR) at ($(S3.east)+(0.5,0)$);
  \draw[dashed, gray!55, line width=0.7pt] ($(sepL)+(0,-4.35)$) -- ($(sepL)+(0,3.85)$);
  \draw[dashed, gray!55, line width=0.7pt] ($(sepR)+(0,-4.35)$) -- ($(sepR)+(0,3.85)$);
\end{scope}
\node[font=\bfseries, align=center] (outTrain) at ($(sepR)+(2.5,0)$)
  {Evolved Skill Banks\\$\mathcal{S}_o^{*},\,\mathcal{S}_w^{*}$};
\draw[line width=1.4pt] ([xshift=0.3cm]mo.north) |- (6.0, 3.85);
\draw[line width=1.4pt] (B.north) |- (6.0, 3.85);
\draw[mainflow] (6.0, 3.85) -| (outTrain.north);

\node[font=\bfseries\small, text=black] at (in.south |- 0,-4.1) {Input};
\node[font=\bfseries\small, text=black] at ($(sepL)!0.5!(sepR) + (0,-4.1)$) {Training phase};
\node[font=\bfseries\small, text=black] at (outTrain.south |- 0,-4.1) {Output};

\end{tikzpicture}%
}
\caption{\textbf{Training (self-evolving)} flow of \textbf{Web2BigTable} over one episode $k$. For each training query $q_k$, Stage~1 reads the long-term orchestrator skills $\mathcal{S}_o$ and decomposes $q_k$ into subtasks $\boldsymbol{\tau}$. Stage~2 dispatches these subtasks to $N$ parallel workers, which read execution skills from $\mathcal{S}_w$ and read/write the short-term workboard $m_e$ until convergence. Stage~3 verifies the aggregated output $X_k$ against the gold reference, produces the structured reflection $r_o^{k+1}$, and consolidates it into both $\mathcal{S}_o$ (via $\mathcal{M}_o$) and $\mathcal{S}_w$ (via $\mathcal{M}_w$). Episodes are processed sequentially: the bottom black loop moves from episode $k$ to $k{+}1$ without replanning within an episode. After $K$ episodes, the two skill banks $(\mathcal{S}_o^{*}, \mathcal{S}_w^{*})$ are frozen and returned as the training output, then used unchanged during inference.}
\label{fig:train-flow}
\end{figure*}

\paragraph{Run--Verify--Reflect loop.}
Inside each training episode $k$, the system first performs one inference pass with its current skills (the inference procedure detailed in Section~\ref{sec:inference}), then compares the produced table $X_k$ against the gold reference $X_k^{\text{gold}}$, and finally distils the resulting error report into skill updates through two parallel reflect operators: $\mathcal{M}_o$ over orchestrator skills (Section~\ref{sec:skill-learning}) and $\mathcal{M}_w$ over worker skills (Section~\ref{sec:skill-evolution}). Both updates are monotone appends of human-readable \texttt{SKILL.md} files; the underlying LLMs are never fine-tuned. The full procedure is summarised in Algorithm~\ref{alg:web2bigtable-train} and illustrated in Figure~\ref{fig:train-flow}.

\subsubsection{Orchestrator Skills Evolution}
\label{sec:skill-learning}
The reflect operator $\mathcal{M}_o$ of Equation~\eqref{eq:strategy-memory-update} is realised as a three-stage pipeline: \textit{Run} executes one training episode, \textit{Verify} produces the structured error report $r_o^{k+1}$, and \textit{Reflect} consumes $r_o^{k+1}$ to synthesise new decomposition skills that append to $\mathcal{S}_o^k$. All three stages are orchestrated by LLM-driven subroutines, with no gradient update applied to the underlying models.

\paragraph{Run.} The system processes a batch of training queries through the full pipeline, archiving structured outputs alongside per-worker JSONL trajectories that capture all tool invocations, intermediate reasoning, and temporal metadata.

\paragraph{Verify.} Outputs are systematically evaluated against gold-standard references in three steps. First, each worker's trajectory is compressed into a structured digest by a highly efficient LLM, capturing the applied decomposition strategy, search queries issued, failure points, and coverage statistics. Second, predicted outputs and gold references are parsed into row sets and compared at the cell level using type-specific scoring (exact match, numeric tolerance, URL normalisation, and LLM-based semantic judgement for free text). Third, results are aggregated into a structured error report highlighting missing row categories, low-accuracy columns, and trajectory anomalies such as context truncation or tool timeouts.

\paragraph{Reflect.} The error report is translated into actionable skills through three sequential steps. Training queries are clustered by structural decomposition pattern rather than semantic topic, typically yielding clusters such as \texttt{split-by-entity} and \texttt{split-by-time-period}. For each cluster, a reflection LLM synthesises a generalisable decomposition skill from the observed errors and successful trajectories. To rigorously prevent overfitting, these skills are constrained to use only structural placeholders. Finally, a router skill is generated that maps query characteristics to the appropriate strategy, enabling the orchestrator to process unseen queries autonomously.

All learned skills are persisted as \texttt{SKILL.md} files, rendering them human-readable, editable, and version-controllable. Concretely, the reflection LLM itself performs the memory write: it emits the newly synthesised decomposition and router skills directly as \texttt{SKILL.md} files into the skill directory backing $\mathcal{S}_o^k$, yielding $\mathcal{S}_o^{k+1}$. Prior skills are never overwritten, so updates are monotone additions. Because skills are plain text consumed via in-context learning, no parameter fine-tuning is required, and the orchestrator simply processes refined instructions during subsequent episodes, closing the slow-timescale loop.

\subsubsection{Worker Skills Evolution}
\label{sec:skill-evolution}

The consolidation operator $\mathcal{M}_w$ of Equation~\eqref{eq:skill-bank-update} is realised by a reflection LLM that emits the skill changes accumulated in episode $k$ directly into the shared bank as \texttt{SKILL.md} entries; as with $\mathcal{M}_o$, the update is a monotone append, so $\mathcal{S}_w^{k} \subseteq \mathcal{S}_w^{k+1}$. The runtime changes that $\mathcal{M}_w$ consolidates arise from two mechanisms operating within each episode.

\paragraph{Skill resolution and creation.} When a specific capability is required, the \texttt{SkillResolver} executes a strictly prioritised search: (1) exact-name matching of local skills within the workspace; (2) semantic retrieval across both local and cloud-based catalogues (comprising over 8{,}000 pre-configured skills), employing a hybrid pipeline that integrates BM25 keyword matching with ChromaDB embedding search (BAAI/bge-m3), subsequently fused via Reciprocal Rank Fusion (RRF) and optionally refined by a cross-encoder; and (3) when no suitable match is found, the \texttt{SkillCreator} module leverages the worker's LLM to synthesise a novel skill on demand, as either a function skill (an executable Python script with a CLI entry point, validated via AST parsing) or a knowledge skill (a structured Markdown document with YAML frontmatter). Newly synthesised skills are instantaneously indexed into the local BM25 and embedding stores.

\paragraph{Error-driven self-repair.} Should a skill execution terminate in failure, the worker initiates an autonomous reflection loop. The error trace and the current skill code are supplied to the LLM to synthesise a corrected iteration, subjected to strict AST validation before replacing the original. The same reflection loop supports extending existing skills with new capabilities via \texttt{evolve\_skill}, whilst preserving backwards compatibility.

These mechanisms operate concurrently at runtime. Because the skill repository is globally shared across active workers via a singleton application context, any skill generated or amended by an individual agent is instantaneously propagated to the collective through an automated index refresh.


\begin{algorithm}[!htbp]
\caption{Web2BigTable Training (Self-Evolving)}
\small
\label{alg:web2bigtable-train}
\begin{algorithmic}[1]
\item[\textbf{Input:}] Training set $\{(q_k, X_k^{\text{gold}})\}_{k=0}^{K-1}$ of $K$ queries paired with ground-truth tables; initial skill banks $\mathcal{S}_o^{0}, \mathcal{S}_w^{0}$; hyperparameters $N$ (worker pool size), $T_{\max}$ (max inner steps).
\item[\textbf{Output:}] Evolved Skill Banks $(\mathcal{S}_o^{*}, \mathcal{S}_w^{*})$ (consumed by Algorithm~\ref{alg:web2bigtable-infer} at inference time).
\FOR{episode $k = 0, 1, \dots, K-1$}
  \STATE \COMMENT{\textit{Stage 1 + Stage 2:} run inference with current skill banks}
  \STATE $X_k \gets$ \textsc{Inference}$(q_k;\, \mathcal{S}_o^{k}, \mathcal{S}_w^{k};\, N, T_{\max})$ \COMMENT{Algorithm~\ref{alg:web2bigtable-infer}}
  \STATE
  \STATE \COMMENT{\textit{Stage 3: Evolve}: score against gold, then consolidate into skill banks}
  \STATE Compute utility $U(X_k) \gets \text{Item-F1}(X_k,\, X_k^{\text{gold}})$ \COMMENT{cell-level training objective}
  \STATE Compress trajectories and aggregate error report $\mathcal{E}^k$
  \STATE Generate reflection $r_o^{k+1} \gets \textsc{Reflect}(\mathcal{E}^k)$
  \STATE $\mathcal{S}_o^{k+1} \gets \mathcal{M}_o(\mathcal{S}_o^{k}, r_o^{k+1})$ \COMMENT{Eq.~\eqref{eq:strategy-memory-update}}
  \STATE $\mathcal{S}_w^{k+1} \gets \mathcal{M}_w(\mathcal{S}_w^{k}, r_o^{k+1})$ \COMMENT{Eq.~\eqref{eq:skill-bank-update}}
\ENDFOR
\STATE \textbf{return} $(\mathcal{S}_o^{K}, \mathcal{S}_w^{K})$ \COMMENT{frozen skill banks $(\mathcal{S}_o^{*}, \mathcal{S}_w^{*})$}
\end{algorithmic}
\end{algorithm}

\subsection{Inference Phase: Query-Time Execution}
\label{sec:inference}

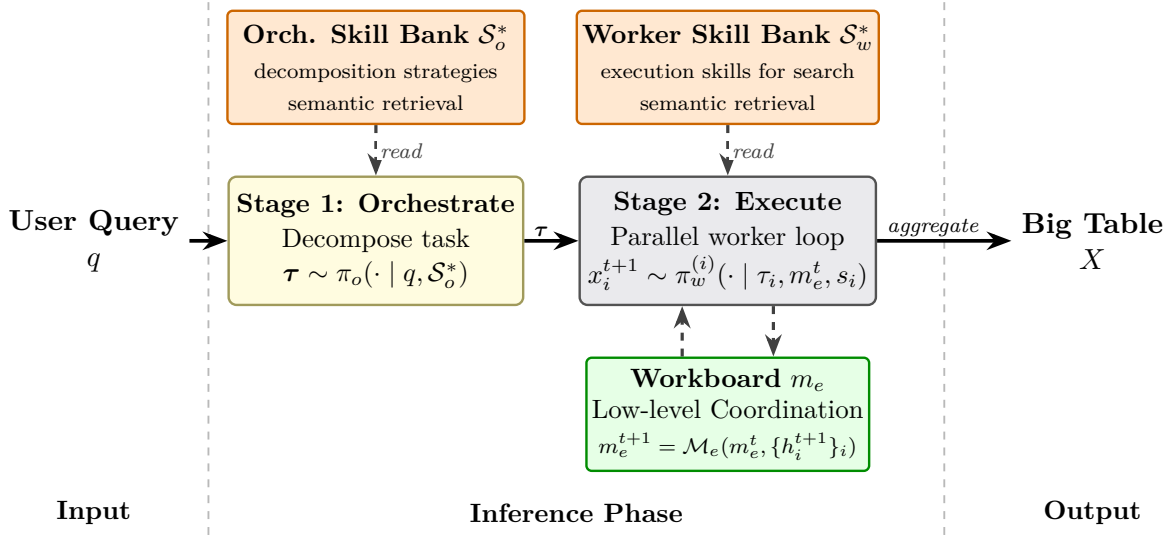
\begin{figure*}[t]
\centering
\begin{tikzpicture}[
>=Stealth, font=\small,
stage/.style={rounded corners=3pt, line width=0.9pt, align=center,
              minimum height=1.7cm, minimum width=3.6cm, inner sep=3pt, draw},
s1/.style={stage, fill=yellow!15,      draw=yellow!55!black},
s2/.style={stage, fill=blue!8!gray!15, draw=gray!65!black},
s3/.style={stage, fill=red!10,         draw=red!55!black},
mem/.style={rounded corners=2pt, line width=0.9pt, align=center,
            minimum height=1.5cm, minimum width=3.4cm, inner sep=2pt, draw},
mem_e/.style={mem, fill=green!10,   draw=green!55!black},
mem_bo/.style={mem, fill=orange!18, draw=orange!80!black},
io/.style={rounded corners=3pt, line width=1.2pt, align=center,
           minimum height=1.2cm, minimum width=2.5cm, inner sep=3pt,
           draw=violet!70!black, fill=violet!12, font=\bfseries},
eq/.style={font=\scriptsize\itshape, fill=white, inner sep=1pt},
mainflow/.style={->, line width=1.4pt},
link/.style={->, line width=1.2pt, black!75, dashed}
]

\node[s1] (S1) at (0.17, 0)
{\textbf{Stage 1: Orchestrate}\\[1pt] Decompose task\\[1pt]
 $\boldsymbol{\tau} \sim \pi_o(\cdot \mid q, \mathcal{S}_o^{*})$};
\node[s2] (S2) at (4.8, 0)
{\textbf{Stage 2: Execute}\\[1pt] Parallel worker loop\\[1pt]
 $x_i^{t+1} \sim \pi_w^{(i)}(\cdot \mid \tau_i, m_e^t, s_i)$};

\node[font=\bfseries, align=center] (out) at (9.6, 0)
{Big Table\\$X$};

\path let \p1=($(S1.east)-(S1.west)$) in
  node[mem_bo, minimum width=\x1] (mo) at (0.17, 2.3)
    {\textbf{Orch. Skill Bank} $\mathcal{S}_o^{*}$\\[1pt]
     {\scriptsize decomposition strategies}\\
     {\scriptsize semantic retrieval}};
\path let \p1=($(S2.east)-(S2.west)$) in
  node[mem_bo, minimum width=\x1] (B) at (4.8, 2.3)
    {\textbf{Worker Skill Bank} $\mathcal{S}_w^{*}$\\[1pt]
     {\scriptsize execution skills for search}\\
     {\scriptsize semantic retrieval}};

\node[mem_e] (me) at (4.8, -2.3)
{\textbf{Workboard} $m_e$\\Low-level Coordination\\[1pt]
 {\scriptsize $m_e^{t+1} = \mathcal{M}_e(m_e^t, \{h_i^{t+1}\}_i)$}};

\node[font=\bfseries, align=center, left=0.5cm of S1] (in) {User Query\\$q$};
\draw[mainflow] (in.east) -- (S1.west);


\draw[mainflow] (S1.east) -- node[eq, above,xshift=-0.13cm]{$\boldsymbol{\tau}$} (S2.west);
\draw[mainflow] (S2.east) -- node[eq, above,xshift=-0.15cm]{aggregate} (out.west);

\draw[link] (mo.south) -- node[eq, midway, right]{read} (S1.north);
\draw[link] (B.south)  -- node[eq, midway, right]{read} (S2.north);

\draw[link] ([xshift=-0.6cm]me.north) -- ([xshift=-0.6cm]S2.south);
\draw[link] ([xshift=0.6cm]S2.south) -- ([xshift=0.6cm]me.north);
\begin{scope}[on background layer]
  \coordinate (sepL) at ($(in.east)!0.5!(S1.west)$);
  \coordinate (sepR) at ($(S2.east)!0.5!(out.west)$);
  \draw[dashed, gray!55, line width=0.7pt] ($(sepL)+(0,-3.9)$) -- ($(sepL)+(0,3.2)$);
  \draw[dashed, gray!55, line width=0.7pt] ($(sepR)+(0,-3.9)$) -- ($(sepR)+(0,3.2)$);
\end{scope}
\node[font=\bfseries\small, text=black] at (in.south       |- 0,-3.6) {Input};
\node[font=\bfseries\small, text=black] at ($(sepL)!0.5!(sepR)+(0,-3.6)$) {Inference Phase};
\node[font=\bfseries\small, text=black] at (out.south      |- 0,-3.6) {Output};

\end{tikzpicture}
\caption{\textbf{Inference} flow of \textbf{Web2BigTable} on an unseen user query $q$. Using the trained skill banks $\mathcal{S}_o^{*}$ and $\mathcal{S}_w^{*}$ as frozen read-only inputs, Stage~1 decomposes $q$ into subtasks $\boldsymbol{\tau}$. Stage~2 runs $N$ parallel workers that resolve execution skills from $\mathcal{S}_w^{*}$ and coordinate through the shared workboard $m_e$ (per-query, short-term); their partial outputs $\{x_i\}$ are aggregated into the structured big table $X$. No verification, reflection, or memory update is performed: the system runs a single forward pass and returns $X$.}
\label{fig:infer-flow}
\end{figure*}

With the skill banks $(\mathcal{S}_o^{*}, \mathcal{S}_w^{*})$ produced by the training phase of Section~\ref{sec:self-evolving}, inference runs a single forward pass over a user query $q$ using these banks as frozen read-only inputs. No reflection, verification, or memory update is performed; the output is a structured table $X$. The procedure is summarised in Algorithm~\ref{alg:web2bigtable-infer} and illustrated in Figure~\ref{fig:infer-flow}.

\subsubsection{Orchestrator}
\label{sec:orchestration}
The orchestration layer governs query decomposition at runtime. Implemented as a single agent powered by a reasoning LLM such as GPT-5 mini, it realises the upper level of the bi-level policy in Equation~\eqref{eq:bilevel-policy}, partitioning the user query $q$ into $N$ self-contained subtasks $\boldsymbol{\tau} = (\tau_1, \dots, \tau_N)$ conditioned on the frozen orchestrator skills $\mathcal{S}_o^{*}$. It first classifies the query by invoking a \texttt{task-router} skill that evaluates structural properties (entity count, schema complexity, temporal requirements, and expected output volume), and then invokes the corresponding decomposition skill. Each resulting subtask specification includes a natural language instruction scoped to a specific data partition, the expected output schema, and a target volume of 10 to 20 items, so that the workload stays strictly within the worker's context limits.

Concurrently, the orchestrator initialises the shared workboard with a subtask checklist and tagged result slots. Once all subtasks have terminated, it performs a validation pass over row counts, column coverage, and data consistency before aggregating worker outputs into a unified table. No post-processing (summarisation, deduplication, etc.) is applied: correctness is expected to emerge directly from the decomposition design.

\subsubsection{Workers}
\label{sec:execution}
The execution layer resolves individual subtasks and coordinates concurrent workers. It realises the lower level of the bi-level policy in Equation~\eqref{eq:bilevel-policy} step by step: at each step $t$, an active worker $i$ samples an action $x_i^{t+1}$ conditioned on its assigned subtask $\tau_i$, the current shared workboard $m_e^t$, and an execution skill $s_i$ retrieved from the frozen worker skills $\mathcal{S}_w^{*}$:
\begin{equation}
  x_i^{t+1} \sim \pi_w^{(i)}\big(\cdot \mid \tau_i, m_e^t, s_i\big).
  \label{eq:worker-policy}
\end{equation}
In direct analogy with Equation~\eqref{eq:single-agent-policy}, each $x_i^{t+1}$ is either a tool call that queries $\mathcal{W}$ (whose observation is appended to the worker's local history $h_i^t$) or a response written to worker $i$'s tagged slot on $m_e$. A worker terminates at the first step $T$ at which $x_i^T$ is a response; we set $x_i := x_i^T$ as worker $i$'s partial output, and the orchestrator aggregates the terminal responses into $X = (x_1, \dots, x_N)$.

\paragraph{Implementation.}
An MCP (Model Context Protocol) server manages the worker pool, dispatching subtasks concurrently via \texttt{asyncio.gather()} with a semaphore controlling the maximum concurrency level (up to 10 workers). Each worker is an independent Memento-Skills~\citep{zhou2026memento} agent powered by Gemini 3 Flash, executing a ReAct loop of reasoning and tool use~\citep{yao2023reactsynergizingreasoningacting}. At each step, the worker either invokes a tool or produces a final response, with generation and tool execution governed by timeouts of 120 seconds and 30 seconds, respectively.

Workers are equipped with eight built-in tools (\texttt{bash}, \texttt{str\_replace}, \texttt{file\_create}, \texttt{view}, \texttt{read\_skill}, \texttt{route\_skill}, \texttt{read\_workboard}, \texttt{edit\_workboard}) alongside dynamically discovered skills. Memory-intensive resources, including the BM25 index, embedding store (BAAI/bge-m3 with ChromaDB), cloud skill catalogue, and cross-encoder reranker, are shared across workers via a singleton application context to eliminate redundant loading. Each worker's complete execution trace is recorded as a JSONL trajectory, providing the foundational input for the self-evolving pipeline.

\paragraph{Shared workboard.}
\label{sec:workboard}
The shared workboard is a structured Markdown document that instantiates the execution memory $m_e$. Unlike message-passing or broadcast architectures, it provides a persistent, globally visible state that all workers can read from and contribute to. It is partitioned into three sections: (i) \textit{Task checklist}: a list of all subtasks and their completion statuses, affording every worker visibility into global progress; (ii) \textit{Worker slots}: a tag-partitioned region where each worker writes results under a unique identifier such as \texttt{<t1\_result>}, preventing write conflicts; and (iii) \textit{Shared context}: background information, constraints, and formatting conventions pre-populated by the orchestrator.

Workers interact with the workboard through two dedicated tools: \texttt{read\_workboard} for observing global progress and peer outputs, and \texttt{edit\_workboard} for appending results to assigned slots. Write operations are protected by file locks and strictly confined to each worker's tagged region, whilst the entire document remains globally readable. The per-step merging of these contributions into $m_e$ follows the update rule $\mathcal{M}_e$ specified in Equation~\eqref{eq:execution-memory-update}.

\paragraph{Dynamic coordination through read-write asymmetry.}
This structural asymmetry is the bedrock of coordination. Because workers can freely observe peer contributions, several adaptive behaviours emerge autonomously:
\begin{itemize}
  \item \textbf{Redundancy avoidance}: a worker observing entities already extracted by a peer skips redundant searches and refocuses on its assigned partition.
  \item \textbf{Coverage gap detection}: by inspecting peer outputs, an active worker can identify missing fields or inconsistencies and dynamically adjust its retrieval strategy.
  \item \textbf{Strategy adaptation}: workers assimilate successful patterns recorded by peers, such as highly effective source URLs, query formulations, or formatting conventions, whilst circumventing approaches that previously resulted in failures.
\end{itemize}

Workers do not merely post final answers. Their LLMs autonomously evaluate what intermediate information warrants sharing, such as experimental plans, partial findings, or discovered URLs, and when to write it. These continuous updates transform the workboard from a static result ledger into a live coordination medium. Because workers execute at heterogeneous speeds, this staggered dynamic creates a natural information cascade, with early-finishing workers enriching the shared context for those still running, progressively refining subsequent outputs without explicit message-passing overhead.

\begin{algorithm}[!htbp]
\caption{Web2BigTable Inference}
\small
\label{alg:web2bigtable-infer}
\begin{algorithmic}[1]
\item[\textbf{Input:}] User query $q$; frozen skill banks $\mathcal{S}_o^{*}, \mathcal{S}_w^{*}$ from Algorithm~\ref{alg:web2bigtable-train}; hyperparameters $N$ (max worker pool), $T_{\max}$ (max inner steps).
\item[\textbf{Output:}] Big Table $X$.
\STATE \COMMENT{\textit{Stage 1: Orchestrate}}
\STATE Classify $q$ via \texttt{task-router} skill in $\mathcal{S}_o^{*}$
\STATE Sample subtasks $\boldsymbol{\tau} = (\tau_1, \dots, \tau_{N_q}) \sim \pi_o(\cdot \mid q, \mathcal{S}_o^{*})$, with $N_q \le N$ \COMMENT{Eq.~\eqref{eq:bilevel-policy}}
\STATE Initialise workboard $m_e^0$ with checklist and tagged slots
\STATE
\STATE \COMMENT{\textit{Stage 2: Execute (asynchronous worker loop)}}
\STATE $t \gets 0$
\WHILE{$\neg \textsc{Converged}(m_e^t)$ \textbf{and} $t < T_{\max}$}
  \STATE Identify active worker set $\mathcal{A}^t \subseteq \{1, \dots, N_q\}$ via asynchronous dispatch
  \FORALL{worker $i \in \mathcal{A}^t$ \textbf{in parallel}}
      \STATE Read current workboard $m_e^t$ via \texttt{read\_workboard}
      \STATE Resolve skill $s_i \gets \textsc{SkillResolver}(\tau_i, \mathcal{S}_w^{*})$
      \STATE Generate sub-output $x_i^{t+1} \sim \pi_w^{(i)}(\cdot \mid \tau_i, m_e^t, s_i)$ \COMMENT{Eq.~\eqref{eq:worker-policy}}
      \STATE \textbf{Lock} slot $\langle \tau_i \rangle$; write $x_i^{t+1}$ via \texttt{edit\_workboard}; \textbf{Unlock}
  \ENDFOR
  \STATE $m_e^{t+1} \gets \mathcal{M}_e(m_e^t, \{h_i^{t+1}\}_{i \in \mathcal{A}^t})$
  \STATE $t \gets t + 1$
\ENDWHILE
\STATE $T \gets t$
\STATE
\STATE \COMMENT{\textit{Aggregate and return}}
\STATE $X \gets \textsc{Aggregate}(\{x_i^{T}\}_{i=1}^{N_q})$, validated against the target table structure implicit in $q$
\STATE \textbf{return} $X$ \COMMENT{big table}
\end{algorithmic}
\end{algorithm}

%% file: sec/experiment.tex
\section{Experiments}
\label{sec:experiments}

We evaluate Web2BigTable across two complementary benchmarks: WideSearch~\cite{wong2025widesearch} for broad-coverage structured extraction, and XBench-DeepSearch~\cite{chen2025xbench} for deep, multi-hop reasoning. We benchmark our framework against state-of-the-art single-agent, end-to-end, and multi-agent systems, alongside comprehensive analysis to isolate the respective contributions of the learned orchestrator strategies, worker-level skill evolution, and workboard-based coordination.

\subsection{Datasets}
\label{sec:dataset}

\paragraph{WideSearch.}
WideSearch~\cite{wong2025widesearch} evaluates the reliability of LLM agents in large-scale, structured information extraction from the live web. It comprises 200 manually curated tasks (100 English, 100 Chinese) spanning 15 domains. Each task requires the agent to extract multi-dimensional atomic data and organise it into a structured table. We adopt a two-phase evaluation protocol. In the \emph{training phase}, we synthesise a set of 20 queries by perturbing task parameters (e.g., entity counts or time ranges) to avoid overlap with the original benchmark, and use these to learn the orchestrator's decomposition skills via the run-verify-reflect pipeline. In the \emph{inference phase}, the strategy memory $m_o$ is frozen and the unmodified 200 tasks serve as the held-out test set.

\paragraph{XBench-DeepSearch.}
XBench-DeepSearch~\cite{chen2025xbench} is a professionally annotated Chinese benchmark designed to assess deep search and tool-use capabilities. Contrasting with WideSearch, which emphasises breadth, XBench-DeepSearch evaluates depth: each task necessitates multi-hop reasoning, cross-source verification, and precise answer extraction from dynamic web content. We adopt the same two-phase protocol. In the \emph{training phase}, we generate 20 modified queries for strategy learning. In the \emph{inference phase}, the strategy memory is frozen and the unmodified dataset serves entirely as the held-out test set.

\subsection{Evaluation Metrics}
\label{sec:metrics}

\paragraph{WideSearch.}
Following the WideSearch evaluation protocol, we report three complementary metrics at increasing levels of granularity:

\begin{itemize}
    \item \textbf{Success Rate (SR):} The most stringent metric, requiring a 100\% match with the ground-truth table. A task is successful only if every row and cell match exactly.
    \item \textbf{Row-level F1:} Treats each table row as a unit, measuring the agent's ability to retrieve complete and correct records. Precision and recall are computed over the matched rows.
    \item \textbf{Item-level F1:} The most granular metric, evaluating individual cells. Cell matching employs type-specific comparators: exact match for categorical fields, numeric tolerance for quantities, URL normalisation for links, and LLM-based semantic judgement for free text.
\end{itemize}

Each system is run four times independently on the test set. We report both \textbf{Avg@4} (mean across four runs) and \textbf{Max@4} (best of four runs) to capture average-case reliability and best-case capability, respectively.

\paragraph{XBench-DeepSearch.}
Following the established benchmark protocol, we report \textbf{Accuracy} evaluated via LLM-as-judge.

\subsection{Model Configurations}
\label{sec:model-config}


\paragraph{Web2BigTable configuration.}
Our model selection balances role-specific capability, cost, and experimental intent. The orchestrator requires planning and reasoning; the workers require fast, reliable tool use. We deliberately pair two lightweight, cost-efficient models with high rate limits to demonstrate that performance stems from framework design rather than backbone capability; stronger LLMs would likely yield further gains (Section~\ref{sec:analysis}). Accordingly, the orchestrator is powered by GPT-5 mini, managing task routing, decomposition, and final synthesis. Each task is served by up to 10 parallel workers powered by Gemini~3 Flash, executing ReAct loops with access to web search, file operations, shell commands, and a shared library comprising over 8,000 cloud skills. Workers coordinate through a markdown-based workboard protected by file locks. The orchestrator's decomposition skills are acquired through the online strategy learning pipeline (Section~\ref{sec:skill-learning}), where each completed training query triggers a three-stage cycle of verification, reflection, and skill updating. On each benchmark, skills are learned from 20 benchmark-specific training queries and frozen prior to test-time evaluation.

\paragraph{Baselines.}
On WideSearch, we compare against three categories of systems:
\begin{enumerate}
    \item \textit{Single Agent}: Frontier LLMs prompted directly with the full task, including Claude Sonnet~4.5, Gemini~3 Pro, GPT-5 High and Doubao-1.6 (with and without thinking mode).
    \item \textit{End-to-End Systems}: Proprietary agentic pipelines from Claude, Gemini, and OpenAI that handle search, extraction, and synthesis as integrated products.
    \item \textit{Multi-Agent Frameworks}: The same frontier LLMs deployed within a standardised orchestration framework, coordinating parallel agents via a hierarchical decomposition pipeline.
\end{enumerate}

On XBench-DeepSearch, baselines include foundation models with tool access (e.g., GLM-4.5, Minimax-M2), proprietary deep research systems (e.g., MiroFlow, OAgents), and open-source agentic models (e.g., DeepMiner-32B-RL, WebShaper-32B). Baseline results are sourced from published reports. For this benchmark, Web2BigTable uses 5 parallel workers rather than 10, as deep search tasks benefit more from sequential depth than parallel breadth.
\subsection{Performance Gain Analysis}
\label{sec:analysis}

We first conduct two sets of analysis experiments. The first isolates the contribution of each system component by disabling individual mechanisms whilst keeping all other configurations identical (Table~\ref{tab:ablation}). The second isolates the contribution of the framework itself from that of the underlying LLMs by comparing Web2BigTable against its own backbone models run as single agents (Table~\ref{tab:framework-vs-llm}). All WideSearch results are reported as Avg@4 over four independent runs; XBench-DeepSearch results are reported as accuracy.

\begin{table}[t]
\centering
\small
\renewcommand{\arraystretch}{1.2}
\begin{tabular}{@{}lcccc@{}}
\toprule
\multirow{2}{*}{Configuration} & \multicolumn{3}{c}{WideSearch (Avg@4)} & XBench \\
\cmidrule(lr){2-4} \cmidrule(lr){5-5}
& SR & Row F1 & Item F1 & Acc. \\
\midrule
\textbf{Full system} & \textbf{38.50} & \textbf{63.53} & \textbf{80.12} & \textbf{73.0} \\
\midrule
w/o learned orch.\ skills & 7.00 & 45.23 & 62.87 & 41.0 \\
w/o workboard & 27.50 & 54.81 & 73.45 & 60.0 \\
w/o worker skill evolution & 33.00 & 59.67 & 76.38 & 64.0 \\
\bottomrule
\end{tabular}
\caption{Contribution of each system component. Removing learned orchestrator skills causes the largest drop across all metrics, confirming that bi-level strategy learning is the primary driver of performance.}
\label{tab:ablation}
\end{table}

\paragraph{Learned orchestrator skills are critical.}
Ablating the learned decomposition skills causes a severe performance drop across both benchmarks. On WideSearch, the Success Rate falls from 38.50 to 7.00, Row F1 from 63.53 to 45.23, and Item F1 from 80.12 to 62.87; on XBench-DeepSearch, accuracy decreases from 73.0 to 41.0, a drop of 32.0 points. Without these skills, the orchestrator defaults to generic LLM-driven decomposition, which induces systematic coverage gaps in WideSearch and disjointed reasoning chains in XBench-DeepSearch. This confirms that slow-timescale orchestrator updates drive the majority of the performance advantage.

\paragraph{Workboard coordination enables dynamic gap recovery.}
Disabling the shared workboard reduces WideSearch Row F1 from 63.53 to 54.81 and XBench-DeepSearch accuracy from 73.0 to 60.0. Without shared memory, workers operate in isolation: they cannot disseminate high-quality sources, perform peer inspection, or facilitate orchestrator-triggered follow-up iterations. On WideSearch, this manifests as redundant queries and unrectified coverage gaps; on XBench-DeepSearch, intermediate findings from preliminary reasoning hops remain inaccessible to downstream agents.

\paragraph{Worker skill evolution provides complementary gains.}
Ablating skill evolution induces a more modest decline, reducing WideSearch Row F1 to 59.67 and XBench-DeepSearch accuracy to 64.0. Constrained to their default toolset, workers lose the capacity to discover task-specific cloud skills or autonomously repair execution failures. The comparatively muted impact indicates that baseline capabilities suffice for the majority of tasks, but the persistent gap confirms that worker-level adaptation yields a consistent complementary enhancement.

\begin{table}[t]
\centering
\small
\renewcommand{\arraystretch}{1.2}
\begin{tabular}{@{}lcccc@{}}
\toprule
\multirow{2}{*}{Configuration} & \multicolumn{3}{c}{WideSearch (Avg@4)} & XBench \\
\cmidrule(lr){2-4} \cmidrule(lr){5-5}
& SR & Row F1 & Item F1 & Acc. \\
\midrule
GPT-5 mini (single agent) & 4.00 & 22.10 & 33.28 & 35.0 \\
Gemini 3 Flash (single agent) & 3.00 & 18.70 & 31.61 & 28.0 \\
\midrule
\textbf{Web2BigTable} {\scriptsize GPT-5 mini + Gemini 3 Flash} & \textbf{38.50} & \textbf{63.53} & \textbf{80.12} & \textbf{73.0} \\
\midrule
$\Delta$ vs GPT-5 mini & +34.50 & +41.43 & +46.84 & +38.0 \\
$\Delta$ vs Gemini 3 Flash & +35.50 & +44.83 & +48.51 & +45.0 \\
\bottomrule
\end{tabular}
\caption{Framework contribution vs underlying LLM capability. The same models that power Web2BigTable achieve far lower scores as single agents, confirming that the performance advantage stems from the framework design rather than model capability.}
\label{tab:framework-vs-llm}
\end{table}

\paragraph{Performance stems from the framework, not the underlying LLMs.}
A natural question is whether Web2BigTable's gains are attributable to the framework or simply to the choice of underlying LLMs. Table~\ref{tab:framework-vs-llm} addresses this directly. GPT-5 mini and Gemini 3 Flash, the two models powering our system, achieve only 33.28 and 31.61 Item F1 respectively as single agents. The full framework reaches 80.12 using these same models, a gain of over 46 points. Crucially, the strongest single-agent baselines in Table~\ref{tab:widesearch} use strictly more powerful models (Claude-4.5-Sonnet at 65.70 Item F1, GPT-5 High at 62.20) yet still fall short of Web2BigTable by at least 14 points. This confirms that the performance advantage stems from bi-level strategy learning, workboard coordination, and runtime skill evolution, not from the raw capability of the backbone LLMs.
\subsection{Benchmark Comparison}
\label{sec:main-results}
\begin{table*}[!ht]
\centering
\small
\setlength{\tabcolsep}{12pt}
\renewcommand{\arraystretch}{1.2}
\resizebox{\textwidth}{!}{%
\begin{tabular}{l*{6}{c}}
\toprule
\multirow{2}{*}{Model / System}
& \multicolumn{2}{c}{Success Rate}
& \multicolumn{2}{c}{Row F1}
& \multicolumn{2}{c}{Item F1} \\
\cmidrule(lr){2-3}
\cmidrule(lr){4-5}
\cmidrule(lr){6-7}
& Avg@4 & Max@4
& Avg@4 & Max@4
& Avg@4 & Max@4 \\
\midrule
\multicolumn{7}{l}{\textit{\textbf{Single Agent}}} \\
\midrule
Claude Sonnet 4 (Thinking) & 2.30 & 5.00 & 31.70 & 41.90 & 57.90 & 66.70 \\
Claude-4.5-Sonnet          & --   & --   & --    & --    & \underline{65.70} & --   \\
Gemini 2.5 Pro             & 1.50 & 5.00 & 30.00 & 41.40 & 51.00 & 63.60 \\
Gemini-3-Pro               & --   & --   & --    & --    & 57.00 & --   \\
OpenAI o3-high             & 4.50 & 9.00 & 34.00 & 44.10 & 52.60 & 62.30 \\
GPT-5 High                 & --   & --   & --    & --    & 62.20 & --   \\
GPT-5 mini                 & 4.00 & 7.50 & 22.10 & 32.40 & 33.28 & 44.10 \\
Kimi K2                         & 1.10 & 3.50 & 29.70 & 41.40 & 54.40 & 65.10 \\
Seed1.8                    & --   & --   & --    & --    & 63.80 & --   \\
DeepSeek-R1                & 0.40 & 1.50 & 20.70 & 31.70 & 41.30 & 55.10 \\
Doubao-1.6                 & 2.60 & 5.00 & 30.00 & 44.10 & 48.30 & 63.90 \\
Doubao-1.6-non-thinking    & 1.00 & 3.50 & 27.20 & 39.90 & 49.00 & 62.00 \\
Gemini 3 Flash             & 3.00 & 6.00 & 18.70 & 28.50 & 31.61 & 42.30 \\
\midrule
\multicolumn{7}{l}{\textit{\textbf{End-to-End System}}} \\
\midrule
Claude      & 2.50 & 5.00 & 24.10 & 33.50 & 48.40 & 58.50 \\
Gemini      & 4.30 & 8.00 & 36.60 & 45.40 & 59.10 & 67.20 \\
OpenAI o3   & 3.00 & 5.50 & 23.90 & 36.00 & 45.50 & 56.50 \\
\midrule
\multicolumn{7}{l}{\textit{\textbf{Multi-Agent Framework}}} \\
\midrule
Claude Sonnet 4 (Thinking) & 3.60 & 6.50 & \underline{38.50} & \underline{52.20} & 62.20 & \underline{73.10} \\
Gemini 2.5 Pro             & 2.00 & 6.50 & 33.50 & 44.60 & 57.40 & 66.30 \\
OpenAI o3-high             & \underline{5.10} & \underline{9.50} & 37.80 & 50.50 & 57.30 & 68.90 \\
Kimi K2                         & 3.00 & 6.50 & 36.20 & 49.60 & 61.20 & 70.70 \\
DeepSeek-R1                & 0.80 & 3.00 & 22.90 & 36.60 & 44.30 & 60.30 \\
Doubao-1.6                 & 2.50 & 5.50 & 34.00 & 48.90 & 54.60 & 69.70 \\
Doubao-1.6-non-thinking    & 2.10 & 4.50 & 29.70 & 42.70 & 52.80 & 65.10 \\
\midrule
\rowcolor{blue!8}
\textbf{Web2BigTable (Ours)} {\scriptsize GPT-5 mini + Gemini 3 Flash} & \textbf{38.50}$_{\color{teal}{+33.40}}$ & \textbf{40.00}$_{\color{teal}{+30.50}}$ & \textbf{63.53}$_{\color{teal}{+25.03}}$ & \textbf{65.12}$_{\color{teal}{+12.92}}$ & \textbf{80.12}$_{\color{teal}{+14.42}}$ & \textbf{82.48}$_{\color{teal}{+9.38}}$ \\
\bottomrule
\end{tabular}
}
\caption{Performance comparison on the WideSearch benchmark. Full results in Appendix~\ref{tab:widesearch_all}.}
\label{tab:widesearch}
\end{table*}

\definecolor{acblue}{RGB}{31,78,121}
\definecolor{accoral}{RGB}{192,75,48}
\definecolor{acteal}{RGB}{29,158,117}
\definecolor{acamber}{RGB}{186,117,23}
\definecolor{acpurple}{RGB}{100,74,166}
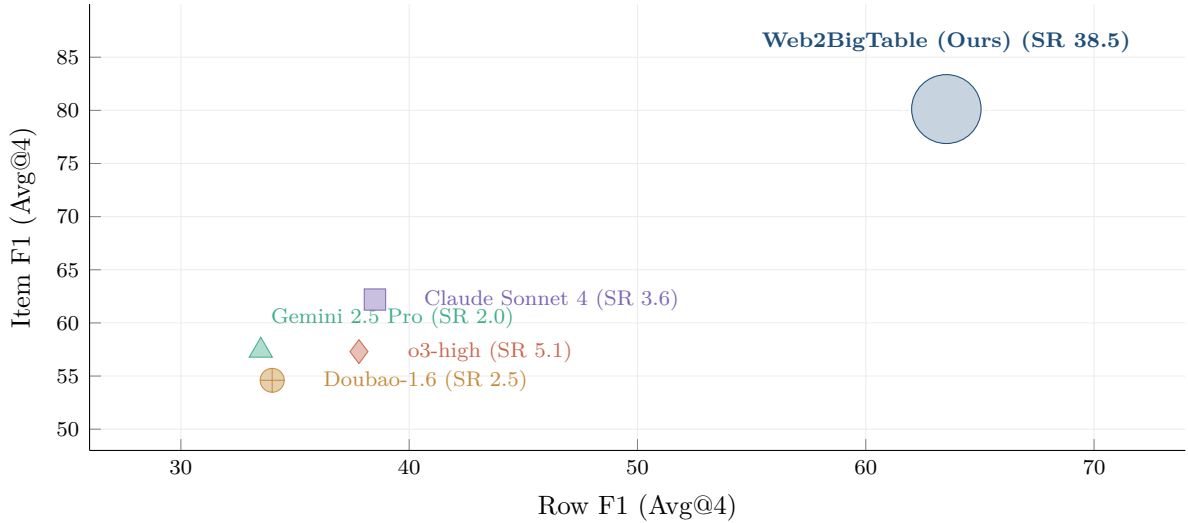
\begin{figure}[t]
\centering
\begin{tikzpicture}
\begin{axis}[
    width=\columnwidth,
    height=7.5cm,
    xlabel={Row F1 (Avg@4)},
    ylabel={Item F1 (Avg@4)},
    xlabel style={font=\small},
    ylabel style={font=\small},
    xmin=26, xmax=74,
    ymin=48, ymax=90,
    xtick={30,40,50,60,70},
    ytick={50,55,60,65,70,75,80,85},
    x tick label style={font=\scriptsize},
    y tick label style={font=\scriptsize},
    axis lines*=left,
    grid=major,
    grid style={line width=0.2pt, draw=gray!15},
    clip=false,
]
\addplot[only marks, mark=diamond*, mark size=4.5pt, draw=accoral!80, fill=accoral!35] coordinates {(37.80, 57.30)};
\addplot[only marks, mark=triangle*, mark size=5pt, draw=acteal!80, fill=acteal!35] coordinates {(33.50, 57.40)};
\addplot[only marks, mark=square*, mark size=4pt, draw=acpurple!80, fill=acpurple!35] coordinates {(38.50, 62.20)};
\addplot[only marks, mark=oplus*, mark size=4.5pt, draw=acamber!80, fill=acamber!35] coordinates {(34.00, 54.60)};
\addplot[only marks, mark=*, mark size=13pt, draw=acblue, fill=acblue!25] coordinates {(63.53, 80.12)};
\node[font=\tiny\bfseries, anchor=center, text=white] at (axis cs:63.53, 80.12) {(SR 38.5)};
\node[font=\scriptsize\bfseries, anchor=south, text=acblue] at (axis cs:63.53, 84.5) {Web2BigTable (Ours) (SR 38.5)};
\node[font=\scriptsize, anchor=west, text=accoral!85] at (axis cs:39.5, 57.30) {o3-high (SR 5.1)};
\node[font=\scriptsize, anchor=south west, text=acteal!85] at (axis cs:33.50, 58.5) {Gemini 2.5 Pro (SR 2.0)};
\node[font=\scriptsize, anchor=west, text=acpurple!85] at (axis cs:40.2, 62.20) {Claude Sonnet 4 (SR 3.6)};
\node[font=\scriptsize, anchor=west, text=acamber!85] at (axis cs:35.8, 54.60) {Doubao-1.6 (SR 2.5)};
\end{axis}
\end{tikzpicture}
\caption{Performance landscape on WideSearch (Avg@4). Scatter points show multi-agent systems: position encodes Row F1 ($x$) and Item F1 ($y$); interior label encodes Success Rate. Web2BigTable dominates all three metrics. Full results are in Table~\ref{tab:widesearch}.}
\label{fig:widesearch_bubble}
\end{figure}

\begin{table}[t]
\centering
\small
\renewcommand{\arraystretch}{1.15}
\begin{tabularx}{\textwidth}{@{}Xr@{}}
\toprule
Model / System & Accuracy \\
\midrule
\multicolumn{2}{@{}l}{\textit{\textbf{Foundation Models with Tools}}} \\
\midrule
Minimax-M2 & \underline{72.0} \\
DeepSeek-V3.2 & 71.0 \\
GLM-4.5 & 70.0 \\
Minimax-M2 (OpenRouter) & 64.0 \\
DeepSeek-V3.2 (OpenRouter)& 65.0 \\
GLM-4.5 (OpenRouter)& 64.0 \\
Claude-4.5-Sonnet & 66.0 \\
Gemini-2.5-Pro & 56.0 \\
GPT-5 mini & 35.0 \\
Gemini 3 Flash & 28.0 \\
\midrule
\multicolumn{2}{@{}l}{\textit{\textbf{Deep Research System}}} \\
\midrule
MiroFlow (GPT-5) & \underline{72.0} \\
Kimi-Researcher & 69.0 \\
OAgents (Claude-3-7) & 54.5 \\
Gemini DeepResearch & 50.0 \\
OpenAI DeepResearch & 26.6 \\
\midrule
\multicolumn{2}{@{}l}{\textit{\textbf{Open-Source Agentic Model}}} \\
\midrule
DeepMiner-32B-RL & 62.0 \\
WebShaper-32B & 54.6 \\
WebSailor-32B & 53.3 \\
AFM-32B-RL & 52.0 \\
WebDancer-QwQ & 40.0 \\
\midrule
\textbf{Web2BigTable (Ours)} {\scriptsize GPT-5 mini + Minimax-M2 (OpenRouter)} & 67.0\\
\textbf{Web2BigTable (Ours)} {\scriptsize GPT-5 mini + DeepSeek-V3.2 (OpenRouter)} & 69.0\\
\textbf{Web2BigTable (Ours)} {\scriptsize GPT-5 mini + GLM-4.5 (OpenRouter)} & 70.0 \\
\rowcolor{blue!8}
\textbf{Web2BigTable (Ours)} {\scriptsize GPT-5 mini + Gemini 3 Flash} & \textbf{73.0}\\
\bottomrule
\end{tabularx}
\caption{Performance comparison on XBench-DeepSearch.}
\label{tab:xbench}
\end{table}

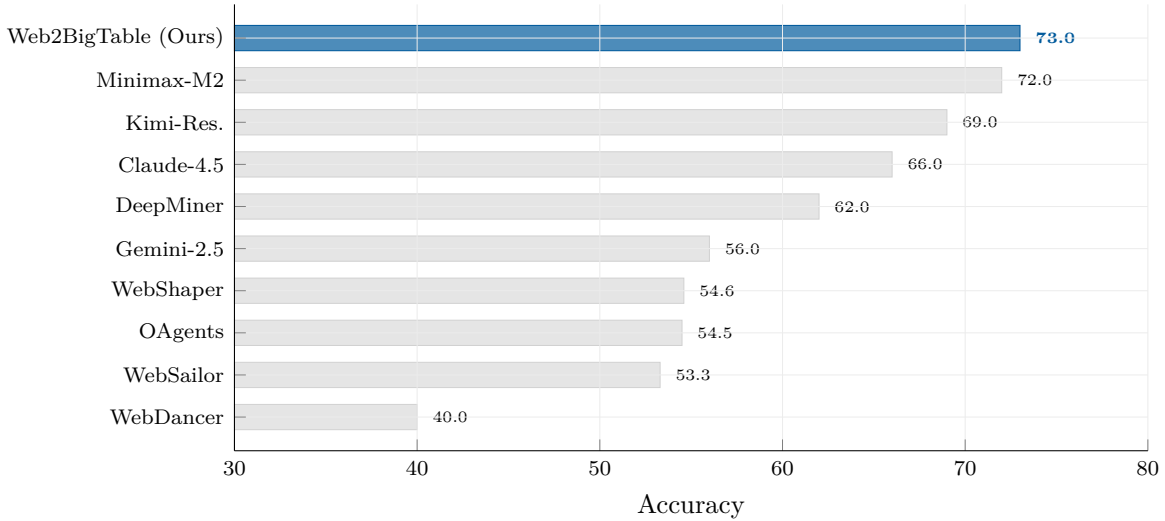
\begin{figure}[t]
\centering
\definecolor{barbase}{RGB}{189,189,189}
\definecolor{baours}{RGB}{0,90,156}
\begin{tikzpicture}
\begin{axis}[
    width=0.85\columnwidth,
    height=7.5cm,
    xlabel={Accuracy},
    xlabel style={font=\small},
    xmin=30, xmax=80,
    xtick={30,40,50,60,70,80},
    ymin=0.2, ymax=10.8,
    ytick={1,2,3,4,5,6,7,8,9,10},
    yticklabels={WebDancer, WebSailor, OAgents, WebShaper, Gemini-2.5, DeepMiner, Claude-4.5, Kimi-Res., Minimax-M2, {Web2BigTable (Ours)}},
    y tick label style={font=\scriptsize},
    x tick label style={font=\scriptsize},
    axis lines*=left,
    clip=false,
    grid=major,
    grid style={line width=0.2pt, draw=gray!15},
    major grid style={line width=0.2pt, draw=gray!15},
    axis on top,
]
\foreach \y/\v in {1/40.0, 2/53.3, 3/54.5, 4/54.6, 5/56.0, 6/62.0, 7/66.0, 8/69.0, 9/72.0} {
    \edef\temp{\noexpand\fill[fill=barbase!40, draw=barbase!70, line width=0.3pt]
        (axis cs:30, \y-0.3) rectangle (axis cs:\v, \y+0.3);
    \noexpand\node[font=\noexpand\tiny, anchor=west, xshift=2pt] at (axis cs:\v, \y) {\v};
    }\temp
}
\fill[fill=baours!70, draw=baours, line width=0.4pt]
    (axis cs:30, 9.7) rectangle (axis cs:73.0, 10.3);
\node[font=\tiny\bfseries, anchor=west, xshift=2pt, text=baours] at (axis cs:73.0, 10) {73.0};
\end{axis}
\end{tikzpicture}
\caption{Accuracy on XBench-DeepSearch.}
\label{fig:xbench_bar}
\end{figure}


\paragraph{WideSearch results.}
Table~\ref{tab:widesearch} and Figure~\ref{fig:widesearch_bubble} present the primary results on WideSearch. Web2BigTable achieves an Avg@4 Success Rate of 38.50 ($7.5\times$ the second best at 5.10), a Row F1 of 63.53 (+25.03 over the second best), and an Item F1 of 80.12 (+14.42 over the second best). As established in Section~\ref{sec:analysis}, these gains are attributable to the framework design rather than backbone model capability. Across all three metric categories, existing multi-agent frameworks plateau below 6 SR, 39 Row F1, and 63 Item F1 despite employing strictly stronger backbone LLMs (o3-high, Claude Sonnet 4). This ceiling arises because static decomposition heuristics induce systematic coverage gaps that no amount of per-worker capability can compensate for. The gains achieved by Web2BigTable are directly attributable to the learned orchestrator skills, which replace these fixed heuristics with task-adaptive decomposition strategies acquired through the run-verify-reflect pipeline.

\paragraph{XBench-DeepSearch results.}
Table~\ref{tab:xbench} and Figure~\ref{fig:xbench_bar} summarise the
XBench-DeepSearch evaluation. Web2BigTable achieves 73.0 accuracy,
surpassing all baselines including frontier proprietary systems such as
Minimax-M2 and MiroFlow (both at 72.0). Rows labelled ``(OpenRouter)''
correspond to our own re-evaluation of the underlying models via the
OpenRouter API. XBench does not disclose its official inference
configuration, including context window, temperature, and decoding
parameters, so the 6 to 8 point gap between these rows and the
officially reported scores (for example, 64.0 versus 72.0 for
Minimax-M2) most likely reflects undocumented setup differences rather
than capability gaps in the underlying models. We therefore treat the
officially reported numbers as the primary point of comparison. The
impact of the learned orchestrator skills is evident: accuracy improves
from 41.0 (without learned skills) to 73.0 following strategy learning,
a gain of 32.0 points from 20 synthesised training queries. This gain
mirrors the pattern observed on WideSearch, where the same
run-verify-reflect pipeline drives the majority of the performance
advantage. The consistency of this effect across two structurally
distinct benchmarks, one emphasising breadth over hundreds of entities
and the other depth over multi-hop reasoning chains, validates the
generalisability of the bi-level framework.

\paragraph{Case Study.}

To illustrate how learned decomposition strategies produce qualitatively different behaviour, we present task \texttt{ws\_en\_006}, a query requiring 534 ground-truth rows across 5 columns spanning 6 Taylor Swift concert tours over 15 years.

\begin{tcolorbox}[casebox=yellow!5, title=1. User Query (\texttt{ws\_en\_006})]
\small
List every concert on Taylor Swift's official tours from Jan 1, 2010 to May 1, 2025. Columns: \textit{Date, Concert Name, Host Country, Host City, Host Venue}. Each show on its own row, in chronological order, no omissions.
\end{tcolorbox}

\begin{tcolorbox}[casebox=red!5, title=2. Baseline Decomposition (without learned orchestrator skills)]
\small
\textbf{Strategy: split-by-time-period} (default LLM reasoning)

\vspace{0.3em}
\begin{tabular}{@{}rl@{}}
Worker 0: & 2010--2011 (all tours mixed) \\
Worker 1: & 2012--2013 \\
Worker 2: & 2014--2015 \\
Worker 3: & 2016--2018 \\
Worker 4: & 2019--2023 $\leftarrow$ \textit{spans 5 years; Eras Tour alone has 130+ dates} \\
Worker 5: & 2024--2025 \\
\end{tabular}

\vspace{0.3em}
\textbf{Failure mode:} Workers handle multiple tours mixed within each time window. Worker 4 is overwhelmed by the 5-year scope. Querying ``Taylor Swift concerts 2019--2023'' returns noisy, unfocused results. \textbf{Result: 234 / 534 rows retrieved.}
\end{tcolorbox}

\begin{tcolorbox}[casebox=green!5, title=3. Learned Orchestrator Skill (self-evolved in Phase 1)]
\small
The task-router skill classified this query as \texttt{split-by-entity}. The decompose skill (\texttt{decompose-split-by-entity/SKILL.md}), automatically generated from training-time error analysis:

\vspace{0.3em}
\begin{tcolorbox}[colback=white, colframe=green!40, boxrule=0.4pt, arc=1pt, left=4pt, right=4pt, top=2pt, bottom=2pt, fontupper=\ttfamily\scriptsize]
\# Split-by-entity decomposition\\
When the query targets a LIST OF NAMED ENTITIES (e.g., tours,\\
brands, athletes), split by entity name --- NOT by time period.\\
Each worker gets one entity as their search keyword.\\[2pt]
Rules learned from past failures:\\
- If any entity has >80 expected items, split further by region.\\
- Always include a gap-detection worker.\\
- If >10\% missing after gap-detection, trigger Round 2.
\end{tcolorbox}
\end{tcolorbox}

\begin{tcolorbox}[casebox=blue!5, title=4. Web2BigTable Decomposition (applying learned skills)]
\small
\textbf{Strategy: split-by-entity (tour name) + split-by-region for large tours}

\vspace{0.3em}
\begin{tabular}{@{}rl@{}}
Worker 0: & Fearless Tour (2010) \\
Worker 1: & Speak Now World Tour (2011--2012) \\
Worker 2: & The Red Tour (2013--2014) \\
Worker 3: & The 1989 World Tour (2015) \\
Worker 4: & Reputation Stadium Tour (2018) \\
Worker 5: & The Eras Tour: North America + early 2023 \\
Worker 6: & The Eras Tour: International 2023--2024 \\
Worker 7: & Gap detection and completeness check \\
\end{tabular}

\vspace{0.3em}
\textit{Round 2:} System detected incomplete Red Tour data $\to$ dispatched 2 additional verification workers.

\vspace{0.3em}
\textbf{Result: 653 raw rows retrieved} across 6 tours; after deduplication and aggregation, \textbf{556 unique rows submitted}. Row F1 = \textbf{93.8\%.}
\end{tcolorbox}

\begin{tcolorbox}[casebox=teal!5, title=5. Per-Tour Retrieval Distribution (after deduplication)]
\small
\centering
\begin{tabular}{@{}lclc@{}}
\toprule
Tour & Rows & Tour & Rows \\
\midrule
The Red Tour (2013--2014) & 142 & Fearless Tour (2010) & 85 \\
The Eras Tour (2023--2024) & 118 & The 1989 World Tour (2015) & 73 \\
Speak Now World Tour (2011--2012) & 93 & Reputation Stadium Tour (2018) & 45 \\
\midrule
\multicolumn{3}{@{}l}{\textbf{Total (deduplicated)}} & \textbf{556} \\
\bottomrule
\end{tabular}
\end{tcolorbox}

\begin{tcolorbox}[casebox=purple!5, title=6. Performance Comparison]
\small
\centering
\renewcommand{\arraystretch}{1.15}
\begin{tabular}{@{}lccc@{}}
\toprule
& \textbf{Row F1} & \textbf{Item F1} & \textbf{Rows} \\
\midrule
GPT-5 mini (single agent) & 15.2\% & 71.3\% & 81 \\
Gemini 3 Flash (single agent) & 12.8\% & 68.5\% & 64 \\
\midrule
Web2BigTable (w/o skills) & 26.8\% & 79.7\% & 234 \\
\textbf{Web2BigTable (full)} & \textbf{93.8\%} & \textbf{96.2\%} & \textbf{556} \\
\midrule
$\Delta$ (full vs best single agent) & +78.6 & +24.9 & +475 \\
$\Delta$ (full vs w/o skills) & +67.0 & +16.5 & +322 \\
\bottomrule
\end{tabular}

\vspace{0.3em}
\raggedright
{\scriptsize \textbf{Key insight:} GPT-5 mini and Gemini 3 Flash each retrieve fewer than 100 rows as single agents. The same two models inside Web2BigTable retrieve 556 deduplicated rows at 93.8\% Row F1, confirming that the gain stems from the framework, not the underlying LLMs.}
\end{tcolorbox}

\captionof{figure}{Case study on task \texttt{ws\_en\_006} (Taylor Swift concerts, 534 ground-truth rows, 6 tours). Workers collectively retrieve 653 raw rows; after deduplication and aggregation, 556 unique rows are submitted for evaluation. The auto-generated orchestrator skill selects entity-based decomposition with adaptive region splitting, achieving 93.8\% Row F1 vs.\ 12.8\%--26.8\% for single-agent and skill-less baselines.}
\label{fig:case_study_a}

%% file: sec/related.tex
\section{Related Work}
\label{sec:related}



\paragraph{Autonomous Web Search and Deep Research Agents}

Early LLM web search focused on single-turn retrieval to mitigate hallucinations (e.g., WebGPT \cite{nakano2021webgpt}, WebGLM \cite{liu2023webglm}). The paradigm subsequently shifted towards autonomous, multi-step web navigation, catalysed by benchmarks like WebArena \cite{zhou2023webarena} and Mind2Web \cite{deng2023mind2web}. Recently, this trajectory culminated in deep research systems designed for exhaustive, long-horizon investigations. Works like WebThinker \cite{li2025webthinker} and Search-R1 \cite{jin2025search} employ reinforcement learning for this purpose, while proprietary frameworks have demonstrated impressive proficiency in deep, multi-hop reasoning \cite{chen2025xbench}. However, while monolithic deep research agents \cite{huang2025deep} excel at vertical reasoning, they face severe limitations in large-scale structured extraction often termed widesearch \cite{wong2025widesearch}. When retrieving numerous entities across heterogeneous sources, single-agent architectures are inevitably bottlenecked by context saturation, error propagation, and rigid task decomposition. This highlights the necessity for scalable frameworks that can parallelise broad extraction without sacrificing reasoning depth.

While emerging systems attempt to tackle these broad web-extraction tasks, most still rely on static heuristics or computationally expensive gradient updates. \textbf{Web2BigTable} diverges by autonomously evolving its web-search and task-decomposition strategies through a closed-loop verify-reflect cycle, entirely without parameter updates. Furthermore, the architecture dynamically discovers, synthesizes, and refines executable search skills at inference time. This provides a highly flexible, training-free alternative that overcomes static reasoning bottlenecks, seamlessly scaling to the extreme breadth of open-web information extraction.

\paragraph{Self-Evolving Agents}

An expanding body of literature investigates how LLM agents progressively enhance their capabilities through experiential learning without parameter updates~\cite{gao2025survey}. Frameworks such as SAMULE~\cite{ge2025samule} and EvolveR~\cite{wu2025evolver} distil transferable insights through multi-level reflection and closed-loop self-distillation. Concurrently, reinforcement learning is increasingly employed to construct and exploit structured skill libraries from sequential rollouts, as demonstrated by SAGE~\cite{sage2025} and SkillRL~\cite{skillrl2026}. Other approaches facilitate continual learning via artefact-centric discovery loops~\cite{belle2025agents} or by decoupling reasoning from learning using hierarchical procedural memory for compositional generalisation~\cite{forouzandeh2025macla}.

Whilst \textbf{Web2BigTable} builds upon this foundation, it distinguishes itself by operating within a \emph{multi-agent} paradigm. Specifically, the central orchestrator evolves macro-level \emph{decomposition strategies}, whilst the parallel worker agents independently cultivate micro-level \emph{execution skills}. Crucially, both evolutionary processes occur concurrently within a unified learning loop, remaining devoid of gradient updates to the underlying language models.

\paragraph{Agentic Memory Systems}
\label{sec:rw-memory}

Memory serves as a foundational component distinguishing autonomous LLM agents from stateless inference~\cite{zhang2025memory_survey, mas_memory_survey2025}. Recent architectures employ sophisticated memory management, such as A-MEM~\cite{amem2025}, which organises memories into interconnected knowledge networks. To manage such structures, Memory-R1~\cite{memoryr1_2025} utilises reinforcement learning to train dedicated managers for structured long-term memory operations, whilst MUSE~\cite{yang2025learning} categorises experiential data via a hierarchical Plan-Execute-Reflect-Memorise loop. Extending these concepts to multi-agent collectives, frameworks like G-Memory~\cite{zhang2025gmemory} trace hierarchical memory across agents, enabling shared episodic-to-semantic consolidation through structured knowledge graphs.

Establishing a rigorous mathematical framework for such systems, the Memento series~\cite{wang2025memento} formalises memory-augmented agents through the Stateful Reflective Decision Process (SRDP), providing convergence guarantees for retrieval policies operating over an evolving memory bank. \textbf{Web2BigTable} inherits this theoretical foundation, extending it to a hierarchical multi-agent architecture. Within this paradigm, skill memories are maintained at two distinct strata: macro-level decomposition strategies for the orchestrator, and micro-level executable skills for the workers. Both tiers are continuously refined via a unified Read-Write Reflective Learning loop.

%% file: sec/conclusion.tex
\section{Conclusion}
\label{sec:conclusion}

We presented Web2BigTable, a bi-level multi-agent framework for large-scale web-to-table construction. The system addresses the fundamental tension between breadth and reliability in agentic web search through a memory-mediated self-evolving architecture: an upper-level orchestrator that automatically learns reusable decomposition strategies from a small training split, and a lower-level pool of asynchronous workers that coordinate through a shared Markdown workboard whilst evolving their own execution skills, with all adaptation mediated through persistent, human-readable memory rather than gradient updates. Empirically, Web2BigTable achieves state-of-the-art performance on the WideSearch benchmark and generalises to depth-oriented search on XBench-DeepSearch, with ablation studies confirming that the learned decomposition skills account for the largest share of the performance advantage. These results demonstrate that bi-level memory-mediated coordination, coupled with automated strategy learning, provides a scalable and training-free alternative to monolithic agent architectures for large-scale structured extraction.




%% file: sec/appendix.tex
\section{Theoretical Extension: Memento-Team}
\label{app:memento-team}
 
The bilevel game formulation presented in this work provides an intuitive framework for understanding the dual-memory dynamics of Web2BigTable. In our ongoing work,  \textbf{Memento-Team: Game-Theoretic Multi-Agent LLM Systems by Bi-level Coordinated Reflection}, we develop a rigorous theoretical generalisation of this framework. Memento-Team formalises the orchestrator--worker interaction as a full Stackelberg game with frozen LLM agents, where the strategy memory and execution memory serve as the sole decision variables. At the lower level, the workers' subgame is further characterised as a discrete exact potential game: under additive utility decomposition, any worker's unilateral memory update that improves its local utility is guaranteed to improve the global objective, ensuring that asynchronous updates on the shared workboard converge without conflict. Building on this formulation, we introduce \textbf{Stochastic Reflective Memory Ascent (SRMA)}, a framework that connects the discrete process of natural language reflection to continuous stochastic optimisation over memory spaces. Under assumptions of bounded communication delay and sparse memory contention, we establish almost sure convergence of the multi-agent system to a neighbourhood of a bilevel memory equilibrium, with the convergence radius governed by the intrinsic hallucination noise of the underlying LLMs. Memento-Team generalises beyond the wide search setting, providing convergence guarantees and design principles applicable to arbitrary multi-agent LLM systems that coordinate through shared non-parametric memory. We plan to validate this generalised framework across a wider spectrum of scenarios, ranging from simple synthetic tasks to complex, practical applications.

\section{Detailed Results}

\begin{table*}[t]
\centering
\caption{Detailed experiments results on the WideSearch benchmark.}
\label{tab:widesearch_all}
\resizebox{1\linewidth}{!}{
\begin{tabular}{l cc cc cc}
\toprule
\textbf{Model / System} & \multicolumn{2}{c}{\textbf{Success Rate}} & \multicolumn{2}{c}{\textbf{Row F1}} & \multicolumn{2}{c}{\textbf{Item F1}}\\
\cmidrule(lr){2-3} \cmidrule(lr){4-5} \cmidrule(lr){6-7}
& Avg@4 & Max@4 & Avg@4 & Max@4 & Avg@4 & Max@4\\
\midrule
\multicolumn{7}{l}{\textit{Single Agent on WideSearch-zh}} \\
Claude Sonnet 4 (Thinking) & 0.25 & 1.00 & 30.19 & 39.73 & 53.76 & 63.19 \\
Gemini 2.5 Pro & 1.00 & 3.00 & 26.95 & 36.96 & 45.57 & 57.26 \\
OpenAI o3-high & 2.00 & 5.00 & 29.30 & 39.31 & 45.19 & 54.46 \\
K2 & 0.25 & 1.00 & 27.79 & 39.03 & 48.81 & 59.64 \\
DeepSeek-R1-0528 & 0.25 & 1.00 & 18.44 & 28.35 & 33.95 & 47.83 \\
Doubao-1.6 & 1.75 & 4.00 & 29.25 & 42.08 & 43.72 & 58.84 \\
Doubao-1.6-non-thinking & 0.50 & 2.00 & 25.56 & 37.41 & 42.87 & 55.79 \\
\midrule
\multicolumn{7}{l}{\textit{Single Agent on WideSearch-en}} \\
Claude Sonnet 4 (Thinking) & 4.25 & 9.00 & 33.18 & 44.08 & 62.02 & 70.27 \\
Gemini 2.5 Pro & 2.00 & 7.00 & 33.05 & 45.82 & 56.38 & 69.97 \\
OpenAI o3-high & 7.00 & 13.00 & 38.70 & 48.84 & 60.03 & 70.08 \\
K2 & 2.00 & 6.00 & 31.54 & 43.68 & 59.91 & 70.52 \\
DeepSeek-R1-0528 & 0.50 & 2.00 & 22.88 & 35.03 & 48.58 & 62.36 \\
Doubao-1.6 & 3.50 & 6.00 & 30.56 & 46.16 & 52.82 & 68.88 \\
Doubao-1.6-non-thinking & 1.50 & 5.00 & 28.86 & 42.31 & 55.06 & 68.17 \\
\midrule
\multicolumn{7}{l}{\textit{Multi-Agent Framework on WideSearch-zh}} \\
Claude Sonnet 4 (Thinking) & 2.75 & 6.00 & 36.85 & 51.46 & 57.13 & 69.53 \\
Gemini 2.5 Pro & 1.00 & 4.00 & 30.93 & 42.21 & 51.79 & 60.87 \\
OpenAI o3-high & 2.75 & 6.00 & 33.83 & 47.85 & 50.35 & 63.06 \\
K2 & 1.25 & 3.00 & 34.74 & 48.01 & 56.86 & 66.75 \\
DeepSeek-R1-0528 & 0.50 & 2.00 & 21.17 & 35.08 & 37.66 & 53.15 \\
Doubao-1.6 & 2.25 & 6.00 & 32.83 & 47.49 & 48.79 & 64.43 \\
Doubao-1.6-non-thinking & 0.50 & 1.00 & 26.93 & 40.30 & 46.52 & 59.63 \\
\textbf{Web2BigTable (Ours)} & \textbf{31.00} & \textbf{32.00} & \textbf{60.32} & \textbf{62.96} & \textbf{76.01} & \textbf{79.33} \\
\midrule
\multicolumn{7}{l}{\textit{Multi-Agent Framework on WideSearch-en}} \\
Claude Sonnet 4 (Thinking) & 4.50 & 7.00 & 40.13 & 52.91 & 67.21 & 76.72 \\
Gemini 2.5 Pro & 3.00 & 9.00 & 36.00 & 47.06 & 63.06 & 71.75 \\
OpenAI o3-high & 7.50 & 13.00 & 41.78 & 53.20 & 64.27 & 74.80 \\
K2 & 4.75 & 10.00 & 37.71 & 51.20 & 65.44 & 74.68 \\
DeepSeek-R1-0528 & 1.00 & 4.00 & 24.57 & 38.10 & 50.91 & 67.54 \\
Doubao-1.6 & 2.75 & 5.00 & 35.14 & 50.38 & 60.39 & 74.87 \\
Doubao-1.6-non-thinking & 3.75 & 8.00 & 32.38 & 44.99 & 58.97 & 70.62 \\
\textbf{Web2BigTable (Ours)} & \textbf{46.00} & \textbf{48.00} & \textbf{66.74} & \textbf{67.28} & \textbf{84.23} & \textbf{85.63} \\
\midrule
\multicolumn{7}{l}{\textit{End-to-End Systems on WideSearch-zh}} \\
Claude & 0.00 & 0.00 & 20.84 & 28.92 & 43.51 & 52.14 \\
Gemini & 1.50 & 4.00 & 32.32 & 40.52 & 52.92 & 60.44 \\
OpenAI o3 & 3.00 & 5.00 & 27.40 & 38.34 & 46.03 & 56.51 \\
\midrule
\multicolumn{7}{l}{\textit{End-to-End Systems on WideSearch-en}} \\
Claude & 5.00 & 10.00 & 27.39 & 38.07 & 53.29 & 64.81 \\
Gemini & 7.00 & 12.00 & 40.95 & 50.29 & 65.18 & 73.90 \\
OpenAI o3 & 3.00 & 6.00 & 20.42 & 33.72 & 45.02 & 56.47 \\
\bottomrule
\end{tabular}
}
\end{table*}
\section{Case Study}
\label{sec:case-study}

\vspace{1em}
\paragraph{Case B.}
This task requires compiling a comprehensive table of every AMD processor with Zen-based architecture released between 2014 and 2024, with 12 columns (Time, Product Series, Processor Model, Core Architecture, Manufacturing Process, Cores, Threads, Core Frequency, L2 Cache, L3 Cache, Graphics Model, Number of Graphics Cores) and 331 ground-truth rows. The high column count makes Item F1 particularly sensitive to per-cell accuracy.

\begin{tcolorbox}[casebox=yellow!5, title=1. User Query (\texttt{ws\_en\_091})]
\small
List all AMD processors with Zen architecture released from Lisa Su becoming CEO (2014) to 2024 inclusive. Columns: Time, Product Series, Processor Model, Core Architecture, Manufacturing Process (nm), Cores, Threads, Core Frequency (GHz), L2 Cache (MB), L3 Cache (MB), Graphics Model, Number of Graphics Cores. Output ``NA'' if information cannot be found.
\end{tcolorbox}

\begin{tcolorbox}[casebox=red!5, title=2. Baseline Decomposition (without learned orchestrator skills)]
\small
\textbf{Strategy: split-by-time-period} (default LLM reasoning)

\vspace{0.3em}
The orchestrator divided the 10-year span into coarse time windows, mixing product lines (Ryzen Desktop, EPYC Server, Threadripper, Mobile, Embedded, PRO) within each window. Workers queried broad terms such as ``AMD Zen processors 2019--2021'', returning incomplete and inconsistent results across the 12 required columns.

\vspace{0.3em}
\textbf{Result: 137 rows retrieved.} Row F1 = 18\%, Item F1 = 32\%.

\vspace{0.3em}
{\scriptsize \textit{Failure: The 12-column requirement amplifies errors: even rows that are partially retrieved have many NA or incorrect cells across cache sizes, graphics models, and manufacturing process fields.}}
\end{tcolorbox}

\begin{tcolorbox}[casebox=green!5, title=3. Learned Orchestrator Skill (auto-generated in Phase 1)]
\small
The task-router classified this query as \texttt{split-by-category}. The decompose skill partitions by \emph{product line}, not by time:

\vspace{0.3em}
\begin{tcolorbox}[colback=white, colframe=green!40, boxrule=0.4pt, arc=1pt, left=4pt, right=4pt, top=2pt, bottom=2pt, fontupper=\ttfamily\scriptsize]
\# Split-by-category decomposition\\
When the query covers MULTIPLE PRODUCT LINES across a long\\
time span, split by product category --- NOT by year.\\
Each worker specialises in one product line across all years.\\[2pt]
Rules learned from past failures:\\
- Assign one worker per major product line (Desktop, Server,\\
\;\; Mobile, Workstation, Embedded, PRO).\\
- For product lines with >50 SKUs, split further by generation.\\
- Always assign a dedicated worker for cache/GPU spec lookup.
\end{tcolorbox}
\end{tcolorbox}

\begin{tcolorbox}[casebox=blue!5, title=4. Web2BigTable Decomposition (applying learned skills)]
\small
\textbf{Strategy: split-by-category (product line) + split-by-generation for large lines}

\vspace{0.3em}
\begin{tabular}{@{}rl@{}}
Worker 0: & Ryzen Desktop (all generations) \\
Worker 1: & EPYC Server (Zen 1--2) \\
Worker 2: & EPYC Server (Zen 3--5) \\
Worker 3: & Ryzen Mobile (U/H/HS/HX series) \\
Worker 4: & Threadripper + Threadripper PRO \\
Worker 5: & Ryzen PRO (Desktop + Mobile) \\
Worker 6: & Athlon + Embedded series \\
Worker 7: & Spec verification (cache, GPU cores) \\
\end{tabular}

\vspace{0.3em}
\textbf{Result: ${\sim}$350 raw rows retrieved} across 8 product lines; after deduplication, \textbf{${\sim}$334 unique rows submitted}. Row F1 = \textbf{89\%}, Item F1 = \textbf{96\%}.
\end{tcolorbox}

\begin{tcolorbox}[casebox=purple!5, title=5. Performance Comparison]
\small
\centering
\renewcommand{\arraystretch}{1.15}
\begin{tabular}{@{}lccc@{}}
\toprule
& \textbf{Row F1} & \textbf{Item F1} & \textbf{Rows} \\
\midrule
GPT-5 mini (single agent) & 10.5\% & 25.3\% & 48 \\
Gemini 3 Flash (single agent) & 8.2\% & 21.7\% & 35 \\
\midrule
Web2BigTable (w/o skills) & 18\% & 32\% & 137 \\
\textbf{Web2BigTable (full)} & \textbf{89\%} & \textbf{96\%} & \textbf{334} \\
\midrule
$\Delta$ (full vs best single agent) & +78.5 & +70.7 & +286 \\
$\Delta$ (full vs w/o skills) & +71.0 & +64.0 & +197 \\
\bottomrule
\end{tabular}

\vspace{0.3em}
\raggedright
{\scriptsize \textbf{Note:} Single-agent scores are for this specific task only, not the benchmark-wide averages in Table~\ref{tab:widesearch}.}

\vspace{0.3em}
{\scriptsize \textbf{Key insight:} With 12 columns per row, single agents retrieve fewer than 50 rows and fill barely 20\% of cells correctly. The learned product-line decomposition enables each worker to query authoritative sources (e.g., AMD product pages, WikiChip) for its specific category, whilst a dedicated spec-verification worker cross-checks cache and GPU fields, achieving 96\% Item F1.}
\end{tcolorbox}

\captionof{figure}{Case B: task \texttt{ws\_en\_091} (AMD Zen processors, 331 ground-truth rows, 12 columns). Workers collectively retrieve ${\sim}$350 raw rows; after deduplication, ${\sim}$334 unique rows are submitted. Single agents retrieve fewer than 50 rows with Item F1 below 26\%. Web2BigTable applies a learned product-line decomposition with dedicated spec-verification workers, achieving 89\% Row F1 and 96\% Item F1.}
\label{fig:case_study_b}

\paragraph{Case C.}
This task requires compiling research papers from two distinct source organisations (ByteDance Seed team and DeepSeek) over a 30-month window, with cross-source temporal verification against arXiv. The defining challenge is twofold: (i) the two sources publish on entirely separate web platforms with different formats, and (ii) when the same paper is mirrored, the canonical date must be reconciled across publisher pages and arXiv submission records.

\begin{tcolorbox}[casebox=yellow!5, title=1. User Query (\texttt{ws\_zh\_069})]
\small
Compile all large-model-related papers published by the ByteDance Seed team and DeepSeek between 1 January 2023 and 30 June 2025. Search the official websites of both organisations (any paper with Seed-team participation counts). For each paper, include the publication date (yyyy-mm-dd), title, and primary authors. If two records refer to the same paper, the canonical date is the arXiv first-submit timestamp. Output a single Markdown table with columns: Organisation, Publication Date, Paper Title, Authors.
\end{tcolorbox}

\begin{tcolorbox}[casebox=red!5, title=2. Baseline Decomposition (without learned orchestrator skills)]
\small
\textbf{Strategy: split-by-time-period} (default LLM reasoning)

\vspace{0.3em}
The orchestrator divided the 30-month window into five 6-month chunks and dispatched workers without distinguishing source organisations. Workers issued generic queries such as ``ByteDance Seed papers 2024 H1'' and ``DeepSeek papers 2024 H2'', returning fragmented results from third-party aggregators (e.g., arXiv listings) rather than the canonical source pages. The smaller-volume DeepSeek catalogue was diluted across all five workers, and arXiv-vs-publisher date conflicts were never reconciled.

\vspace{0.3em}
\textbf{Result: 67 rows retrieved.} Row F1 = 41\%, Item F1 = 58\%.

\vspace{0.3em}
{\scriptsize \textit{Failure modes: (i) the Seed team's $\sim$120 papers were sparsely sampled because each time-window worker treated Seed as just one of many possible sources; (ii) DeepSeek's smaller catalogue lost coverage entirely in early time windows where it had no releases; (iii) date inconsistencies between Seed homepage listings and arXiv records were left unresolved, contaminating the Publication Date column.}}
\end{tcolorbox}

\begin{tcolorbox}[casebox=green!5, title=3. Learned Orchestrator Skill (auto-generated in Phase 1)]
\small
The task-router classified this query as \texttt{split-by-source}. The decompose skill partitions by \emph{source organisation}, with a dedicated worker for cross-source temporal verification:

\vspace{0.3em}
\begin{tcolorbox}[colback=white, colframe=green!40, boxrule=0.4pt, arc=1pt, left=4pt, right=4pt, top=2pt, bottom=2pt, fontupper=\ttfamily\scriptsize]
\# Split-by-source decomposition\\
When the query enumerates records from MULTIPLE NAMED\\
SOURCES (e.g., specific organisations, repositories, or\\
publishers), assign one worker per source rather than\\
splitting along time or topic.\\[2pt]
Rules learned from past failures:\\
- One worker per named source; never mix sources within\\
\;\; a single worker.\\
- For sources with $>$50 records, split further by year\\
\;\; within that source.\\
- If the query specifies a canonical reconciliation rule\\
\;\; (e.g., ``arXiv date takes precedence''), assign a\\
\;\; dedicated verification worker that cross-references\\
\;\; the field across sources.
\end{tcolorbox}
\end{tcolorbox}

\begin{tcolorbox}[casebox=blue!5, title=4. Web2BigTable Decomposition (applying learned skills)]
\small
\textbf{Strategy: split-by-source + split-by-year for the larger source + dedicated verification worker}

\vspace{0.3em}
\begin{tabular}{@{}rl@{}}
Worker 0: & Seed team: 2023 (full year) \\
Worker 1: & Seed team: 2024 (Jan--Jun) \\
Worker 2: & Seed team: 2024 (Jul--Dec) \\
Worker 3: & Seed team: 2025 (Jan--Jun) \\
Worker 4: & DeepSeek: entire 30-month window \\
Worker 5: & arXiv cross-source date verification \\
Worker 6: & Author-list normalisation and duplicate detection \\
\end{tabular}

\vspace{0.3em}
Round 2: The verification worker flagged 8 rows where the publisher-listed date diverged from arXiv submission by more than 30 days; the orchestrator dispatched a follow-up worker to reconcile these specific entries against arXiv before final aggregation.

\vspace{0.3em}
\textbf{Result: 134 rows retrieved.} Row F1 = \textbf{91\%}, Item F1 = \textbf{94\%}.
\end{tcolorbox}

\begin{tcolorbox}[casebox=purple!5, title=5. Performance Comparison]
\small
\centering
\renewcommand{\arraystretch}{1.15}
\begin{tabular}{@{}lccc@{}}
\toprule
& \textbf{Row F1} & \textbf{Item F1} & \textbf{Rows} \\
\midrule
GPT-5 mini (single agent) & 22.4\% & 39.6\% & 29 \\
Gemini 3 Flash (single agent) & 18.7\% & 34.2\% & 23 \\
\midrule
Web2BigTable (w/o skills) & 41\% & 58\% & 67 \\
\textbf{Web2BigTable (full)} & \textbf{91\%} & \textbf{94\%} & \textbf{134} \\
\midrule
$\Delta$ (full vs best single agent) & +68.6 & +54.4 & +105 \\
$\Delta$ (full vs w/o skills) & +50.0 & +36.0 & +67 \\
\bottomrule
\end{tabular}

\vspace{0.3em}
\raggedright
{\scriptsize \textbf{Key insight:} The two sources have asymmetric catalogue sizes ($\sim$120 vs $\sim$10). Single agents retrieve fewer than 30 rows, missing the majority of Seed papers and most DeepSeek entries entirely. Source-based decomposition assigns dedicated workers per organisation, whilst a verification worker resolves arXiv-vs-publisher date conflicts that all other configurations silently ignore.}
\end{tcolorbox}

\captionof{figure}{Case C: task \texttt{ws\_zh\_069} (LLM papers from ByteDance Seed and DeepSeek, $\sim$130 ground-truth rows across two asymmetric sources). Single agents retrieve fewer than 30 rows with Item F1 below 40\%. Web2BigTable applies a learned source-based decomposition with a dedicated arXiv verification worker, achieving 91\% Row F1 and 94\% Item F1.}
\label{fig:case_study_c}

\clearpage